\documentclass[conference]{IEEEtran}
\IEEEoverridecommandlockouts
\pdfoutput=1
\usepackage{cite}
\usepackage{textcomp}
\usepackage{xcolor}
\usepackage{amsfonts}
\usepackage{soul}
\usepackage{url}
\usepackage[utf8]{inputenc}
\usepackage{graphicx}
\usepackage{amsmath}
\usepackage{booktabs}
\usepackage{algorithm}
\usepackage{algpseudocode}
\usepackage{esvect}
\usepackage{booktabs}
\usepackage{array}
\usepackage{multirow}
\usepackage[export]{adjustbox}
\usepackage{listings}
\usepackage[frozencache,cachedir=.]{minted}
\setminted{fontfamily=zlmtt}
\usepackage{amsmath,amssymb,amsfonts,amsthm}
\usepackage{xcolor}
\usepackage{fancyhdr}
\def\BibTeX{{\rm B\kern-.05em{\sc i\kern-.025em b}\kern-.08em
    T\kern-.1667em\lower.7ex\hbox{E}\kern-.125emX}}
\usepackage{cite}
\usepackage{url}
\usepackage{graphicx}
\usepackage{latexsym}
\usepackage{amsmath}
\usepackage{amsthm}
\usepackage{amssymb}
\usepackage{amsfonts}
\usepackage{comment}
\usepackage{array}
\usepackage{xspace}
\usepackage{psfrag}
\usepackage{subfigure}
\usepackage{color}
\usepackage{indentfirst}
\usepackage{multirow}
\usepackage{amstext}
\usepackage{hyperref}
\usepackage{enumitem}
\usepackage{bbm,multirow}
\usepackage{subfigure,wrapfig,amsmath}
\usepackage{balance}
\usepackage{amsfonts}
\usepackage{float}
\usepackage{amsmath,bm}
\usepackage{algorithm}  
\usepackage{algorithmicx}  
\usepackage{algpseudocode}
\theoremstyle{definition}

\newtheorem{theorem}{Theorem}

\newtheorem{Definition}{Definition}

\newcommand{\eg}{\emph{e.g.}}

\newcommand{\presec}{\vspace{-0.02in}}
\newcommand{\postsec}{\vspace{-0.02in}}
\newcommand{\presub}{\vspace{-0.02in}}
\newcommand{\postsub}{\vspace{-0.02in}}
\def\code#1{\texttt{#1}}

\newcommand{\sysname}{\textsc{AGO}\xspace}{}
\newcommand{\sys}{\textsc{AGO}\xspace}

\newcommand\blfootnote[1]{%
\begingroup\renewcommand\thefootnote{}\footnote{#1}%
  \addtocounter{footnote}{-1}%
  \endgroup
}

\usepackage{hyperref}
\usepackage[capitalize]{cleveref}
\begin{document}
\title{\sys: Boosting Mobile AI Inference Performance by Removing Constraints on Graph Optimization}
\lhead{This paper has been accepted by INFOCOM 2023.}
\author{
    \IEEEauthorblockN{Zhiying Xu$^\ddag$\IEEEauthorrefmark{1}, Hongding Peng$^\ddag$\IEEEauthorrefmark{1}, Wei Wang\IEEEauthorrefmark{2},}
    \IEEEauthorblockA{\IEEEauthorrefmark{1}Nanjing University, China
    \{zyxu, mg20330044\}@smail.nju.edu.cn}\IEEEauthorblockA{\IEEEauthorrefmark{2}Nanjing University, China
    ww@nju.edu.cn}
}
\maketitle
\blfootnote{$^\ddag$ Equal contributions.}
\thispagestyle{empty}
\sloppy
{
\begin{abstract}
Traditional deep learning compilers rely on heuristics for subgraph generation, which impose extra constraints on graph optimization, \eg, each subgraph can only contain at most one complex operator.
In this paper, we propose \sys, a framework for graph optimization with arbitrary structures to boost the inference performance of deep models by removing such constraints.
To create new optimization opportunities for complicated subgraphs, we propose intensive operator fusion, which can effectively stitch multiple complex operators together for better performance.
Further, we design a graph partitioning scheme that allows an arbitrary structure for each subgraph while guaranteeing the acyclic property among all generated subgraphs.
Additionally, to enable efficient performance tuning on complicated subgraphs, we devise a novel divide-and-conquer tuning mechanism to orchestrate different system components.
Through extensive experiments on various neural networks and mobile devices, we show that our system can improve the inference performance by up to $3.3\times$ when compared with state-of-the-art deep compilers.
\end{abstract}
\presec \section{Introduction}\label{sec:intro} \postsec

Deep learning has become an essential building block for various mobile applications, such as machine translation and recommendation systems. 
The user experience of these mobile applications are critically impacted by the efficiency of running deep learning inference tasks on mobile devices.
Therefore, code optimization of tensor operators, such as convolution and matrix multiplication, becomes an important research issue for mobile systems.
While manual optimization can lead to order-of-magnitude reduction in inference delay \cite{alibaba2020mnn}, it typically incurs tremendous human efforts, as the tuning process highly depends on the specific hardware architecture as well as the neural network structure. 
To relieve developers from the burden of hand-tuning, researchers design deep learning compilers, such as XLA \cite{leary2017xla}, TVM \cite{chen2018tvm}, and Tiramisu \cite{baghdadi2019tiramisu}, to perform automatic code optimization with compilation and auto-tuning techniques.
In these compilers, each operator is represented as a node in a computational graph, and the tuning result for an operator is called a \emph{schedule}. 

The system design of deep compilers can be divided into two major layers. 
On the top layer, a \emph{graph frontend} partitions the computational graph into multiple subgraphs. 
In contrast with the huge optimization space of the whole graph, optimizing each subgraph separately is more manageable.
To achieve this, existing graph frontends, such as Relay \cite{roesch2018relay} and Apollo \cite{apollo}, group adjacent operators into the same subgraph according to specific heuristics.
As a result, a subgraph can contain at most one complex operator (\textit{e.g.}, convolution and matrix multiplication) and other simple operators (\textit{e.g.}, padding, add, and ReLU).
%
%
%
The bottom layer is \emph{tuner backend}, where enormous schedules are explored in the tuning space of each subgraph separately. 
In general, the best schedule is composed of the optimal parameters for various code optimization techniques, \textit{e.g.}, selected data layouts, tile sizes for loop tiling, and operator fusion schemes. 
Existing tuners either exploit search-based methods \cite{chen2018learning,zheng2020flextensor,zheng2020ansor} or polyhedral models \cite{akg,apollo} to achieve such exploration. 
After exploration, the compiler will generate an optimized tensor program based on the best schedule.

Unfortunately, existing works cannot handle arbitrary graphs efficiently, especially when facing emerging new neural models that tend to use complex structures \cite{mobilenet, mnasnet, sqeezenet, ma2018shufflenet, DBLP:journals/corr/abs-1908-08962, bhargava2021generalization, mehta2022mobilevit}.
First, the graph frontend heavily relies on offline hard-coded heuristics to perform graph partitioning. 
These heuristics only produce simple subgraph structures while overlooking other optimization opportunities. 
They also do not have enough scalability to cater to new neural networks. 
Second, the tuner implicitly limits itself to a small tuning space, because the space is bounded by the simple subgraph generated by the frontend. 
%
Such strict constraints on graph optimization seriously compromises the inference performance. 
%
For example, existing frontends cannot generate subgraphs with multiple complex operators to enable intensive fusion and joint optimization. 
By contrast, we observe that removing this constraint can achieve up to $3.3\times$ speedup for end-to-end inference.

%

%
%

This paper proposes \sys, a framework that enables \underline{a}rbitrary structure \underline{g}raph \underline{o}ptimization to boost the inference performance of mobile deep learning.
In the top graph frontend, \sys exploits a new weighted clustering algorithm to perform graph partitioning, each of the generated subgraphs is free of prior constraints and may contain multiple complex operators. In the bottom backend, \sys designs a more powerful tuner, which can automatically explore schedules for any subgraph. Additionally, different from prior arts, \sys incorporates an extra middle reformer layer to orchestrate the frontend and the backend for efficient subgraph optimization.
%

We need to address three unique challenges to achieve the arbitrary structure graph optimization.

%
%

\noindent\emph{Challenge 1: How to remove the constraints on subgraph structures while keeping the network acyclic?} 
Allowing arbitrary subgraph structures means that any edge in the original graph can cross a cut in the partition.
However, this can lead to cycles among the generated subgraphs.
Cyclic dependencies will result in deadlocks when executing these subgraphs at the runtime. 
%
%
To address this issue, we analytically scrutinize the \emph{inter-subgraph data dependency} given a directed computational graph.
We then show that we can safely group operators in the \emph{affix set} without generating cycles.
Based on our analysis, we devise an iterative clustering algorithm achieving the acyclic property in the graph frontend.

\noindent\emph{Challenge 2: How to efficiently tune arbitrary subgraphs?} 
The primary hurdle to optimize a subgraph with any structure lies in the case that multiple complex operators reside in the same subgraph, which we call \emph{complicated subgraphs}.
%
Although a complicated subgraph can expand the search space to open more optimization opportunities,
%
%
directly exploring schedules in such a large space will incur formidable computational costs, hence inefficient tuning. 
We use two schemes to improve the tuning efficiency. 
First, when generating subgraphs in the graph frontend, we assign a weight for each operator via systematically modeling the relationship between the subgraph structure and the tuning complexity. 
Therefore, we can easily avoid unreasonably huge subgraphs by suppressing the weight. 
Second, during tuning, we propose a divide-and-conquer mechanism in the reformer layer to handle a complicated subgraph. 
The reformer layer further splits a subgraph into several mini-subgraphs, each of which is small to be tuned efficiently.
After several rounds of tuning mini-subgraphs, the reformer layer will join them back as a large subgraph for further optimization. 

\noindent\emph{Challenge 3: How to create new optimization opportunities given a complicated subgraph?} 
One of the most influential tuning techniques to optimize a subgraph is to fuse operators together so that expensive memory accesses can be reduced.
However, different from the conventional operator fusion, fusing multiple complex operators in a complicated subgraph can induce redundant computation, which poses an enormous challenge for the bottom tuner. 
To avoid the dilemma between the redundancy of fusion and the insufficient optimization without fusion, we systematically analyze the \emph{inter-operator data dependency} in complicated subgraphs.
We then discover two categories of subgraph structures that can enable operator fusion while obviating re-computation, which we call \emph{intensive fusion}. 
Therefore, we can exploit new optimization techniques when a complicated subgraph falls into one of the two categories. 
Additionally, when this condition is unmet, our tuner can still benefit from joint optimization for all operators in a complicated subgraph, while the tuning efficiency is already emphasized by the reformer layer.


In summary, we make the following contributions:

$\bullet$ We reveal that the strict constraints imposed on graph optimization is a major obstacle for deep learning compilers to catering to emerging complicated neural architectures.

$\bullet$ We design a weighted clustering algorithm to perform graph partitioning in the frontend, which removes the constraints on subgraph structures while guaranteeing the acyclic property in the resulting partition.

$\bullet$ We devise a divide-and-conquer tuning mechanism to efficiently tune complicated subgraphs, which serves as a middle layer to orchestrate the frontend and the backend.

$\bullet$ We craft a powerful tuner in the backend, which automatically and effectively optimizes complicated subgraphs through the proposed intensive fusion technique.

We integrate \sys into an existing deep compiler and conduct extensive experiments for evaluation. 
The results show that \sys improves the inference performance by up to $3.3\times$ compared with state-of-the-art hand-tuned libraries and auto-tuning frameworks, \textit{e.g.}, Torch Mobile \cite{paszke2019pytorch} and Ansor \cite{zheng2020ansor}.
\presec \section{System Overview}\label{sec:system}

\begin{figure}
	\centering
	\includegraphics[width=0.35\textwidth]{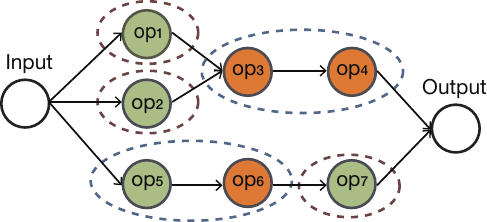}
 	\vspace{-0.1in}
	\caption{An illustrative computational graph.}
	\label{fig:DNNmodel}
 	\vspace{-0.2in}
\end{figure}

\begin{figure*}
	\centering
	\includegraphics[width=0.8\textwidth]{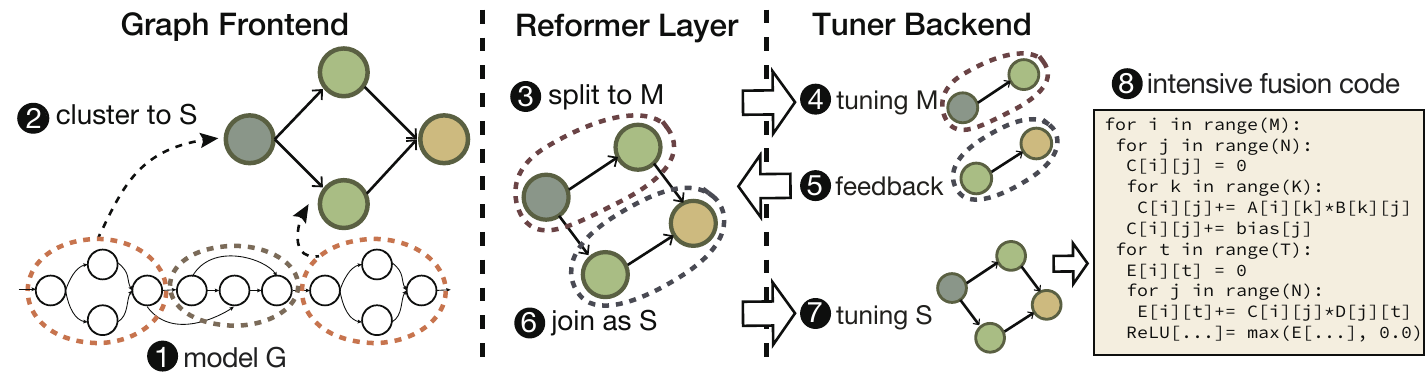}
	\vspace{-0.15in}
	\caption{System overview of \sys.}
	\label{fig:overview}
	\vspace{-0.25in}
\end{figure*}

The input for a deep compiler is a deep learning model, which can be represented as a computational graph where operators and tensors are denoted as nodes and edges respectively, as illustrated in \cref{fig:DNNmodel}.
Complex operators, such as convolution and matrix multiplication operators, are represented as green nodes in \cref{fig:DNNmodel}, while the orange nodes represent simple operators, \textit{e.g.}, add, ReLU, and normalization.
%
%
In prior deep compilers \cite{chen2018tvm, roesch2018relay, zheng2020ansor, apollo}, the computational graph is first partitioned by heuristics into many small subgraphs. 
Each subgraph in these frameworks can contain at most one complex operator. 
%
%
Thus, $op_1$ and $op_2$ must reside in two different subgraphs, although they share the same input tensor and can be stitched together to improve data locality. 
Operators $op_3$ and $op_4$ may constitute another subgraph, even if such no-complex subgraph is trivial, hence no room for performance tuning.
%
The other branch in \cref{fig:DNNmodel} contains two complex operators, $op_5$ and $op_7$, which are forced to be partitioned into two subgraphs although combining all the three operators may benefit from intensive operator fusion.
Therefore, existing heuristics generate unbalanced, small, and simple subgraph structures and hinder further optimization opportunities.

We observe that such inefficient partitioning originates from two aspects. First, the heuristics employed in the graph frontend introduce many unnecessary constraints on the subgraph structure. Second, the underlying tuner backend cannot handle complicated subgraph structures due to the over-simplification of the tuning space. Both of them contribute significantly.

To address this issue,
\sys exploits a weighted clustering algorithm to perform graph partitioning which allows arbitrary subgraph structures. Moreover, \sys designs a more powerful tuner achieving intensive fusion to handle any complicated subgraphs.
For instance, $op_1$, $op_2$, $op_3$, and $op_4$ can be grouped together for operator fusion and joint optimization. Also, we can place $op_5$, $op_6$, and $op_7$ in the same subgraph in the frontend, and then intensively fuse them in the backend to further boost the performance.
%

The workflow of \sys is illustrated in \cref{fig:overview}.
\begin{enumerate}
    \item Given a model file generated by common deep learning frameworks (\textit{e.g.}, TensorFlow \cite{abadi2016tensorflow}), the graph frontend first resolves it into a computational graph $G$.
    \item The frontend partitions operators in $G$ into $n$ subgraphs, each of which is denoted as $S_i$, where $1 \leq i \leq n$.
    \item In the reformer layer, \sys further splits each $S_i$ into $m_i$ mini-subgraphs, each of which is denoted as $M_{ij}$, where $1 \leq j \leq m_i$.
    \item We then offload each $M_{ij}$ as a tuning task for the tuner backend.
    \item After preliminary mini-subgraph optimization, the backend provides the tuned schedules as feedbacks to the reformer layer.
    \item Depending on the feedback, the reformer layer joins the selected mini-subgraphs $M_{ij}$ back as a large subgraph.
    \item Each of the merged subgraph $S_i$ becomes a new tuning task for the tuner backend.
    \item Finally, after optimizing each subgraph $S_i$, we will generate more efficient codes based on the tuned schedules.
\end{enumerate}
Next, we will elaborate our system in a bottom-up manner.
\presec \section{Subgraph Optimization in Backend}\label{sec:fusion}\postsec
In deep models, a tensor operator can be implemented as deeply-nested loops in the source code.
Thus most optimization techniques can be achieved by loop transformation. 
For instance, we can split and reorder loops to achieve \emph{loop tiling}, or merge loops of two operators for \emph{operator fusion}. 
In practice, tiling and fusion are nearly the most two influential techniques for performance optimization.
Specially, operator fusion works on multiple operators, hence is dependent on subgraph structures.
%
%
To optimize arbitrary subgraphs, our major solution in the tuner backend is to craft a new operator fusion scheme, named as \emph{intensive operator fusion}.

In this section, we first introduce how conventional operator fusion works in a mini-subgraph $M_{ij}$. 
Then, we will depict the intensive fusion and how to exploit it for a subgraph $S_i$.

\presub \subsection{Conventional Operator Fusion for Mini-subgraph} \label{sec:normal}
Split from a subgraph by the later reformer layer, each mini-subgraph $M_{ij}$ contains at most one complex operator. 
Suppose $M_{ij}$ contains three operators: 2-d convolution, bias addition, and ReLU, which is a typical workload for recent convolutional neural networks. 
In the following, we will use this mini-subgraph as an example to illustrate how operator fusion improves its computational efficiency.

\begin{figure}
    \centering
    \begin{minted}[fontsize=\footnotesize]{python}
for n in range(N):
 for o in range(O):
  for h in range(H):
   for w in range(W):
    Conv[n,c,h,w] = 0.0
    for ri, rr, rc in range(I, R, C):
     Conv[n,o,h,w] += \ 
      Inp[n,ri,h+rr,w+rc] * Weight[o,ri,rr,rc]
for n, o, h, w in range(N, O, H, W):
  Sum[...] = Conv[...] + Bias[o]
for n, o, h, w in range(N, O, H, W):
  ReLU[...] = max(Sum[...], 0.0)
\end{minted}
\vspace{-0.1in}
    \caption{Loop nest of a mini-subgraph without fusion.}
    \vspace{-0.15in}
    \label{code:mini_F}
\end{figure}

We present the initial loop nest of $M_{ij}$ in \cref{code:mini_F}
\footnote{Direct convolution is preferable than matrix multiplication based implementation \cite{wang2021asymo}. The latter is often used when lacking direct library supports.}
, where $N, O, H, W$ represent the batch size, the number of output channels, the height, and the width of the output tensor, respectively. 
$I$ is the number of input channels, and $R, C$ are the height and the width of the convolutional window. 
Besides, the three reduction loops $I, R, C$ are written in one line for simplicity.
In this program, the bias addition is executed after the whole convolution. 
When the \code{Conv} tensor is large, most of its elements will have been spilled out of the cache at bias addition. 
Subsequently, we must fetch these elements from the main memory into cache again when performing the addition. Such many additional cache misses lead to poor performance, especially on mobile devices with small caches.

\begin{figure}
    \centering
    \begin{minted}[fontsize=\footnotesize]{python}
for n, o, h, w in range(N, O, H, W):
 Conv[n, o, h, w] = 0.0
 for ri, rr, rc in range(I, R, C):
  Conv[...] += Inp[...] * Weight[...]
 Sum[...] = Conv[...] + Bias[o]
 ReLU[...] = max(Sum[...], 0.0)
\end{minted}
\vspace{-0.1in}
    \caption{Conventional fusion within a mini-subgraph.}
\vspace{-0.25in}
    \label{code:loop_fusion_mini}
\end{figure}
We can perform operator fusion within $M_{ij}$ to strike this issue, as illustrated in \cref{code:loop_fusion_mini}. 
Once each element of the \code{Conv} tensor is calculated, the following addition and ReLU operations will be performed. 
Thus, data elements are immediately consumed by downstream operators while still in cache, hence improved operational intensity and inter-operator data locality.
%

%
\vspace{-0.05in}
\presub \subsection{Intensive Operator Fusion for Subgraph} \label{sec:deeper}\postsub
Conventional operator fusion only stitches a complex operator with its following simple operators, hence is also named as epilogue fusion. 
To optimize subgraphs with complicated structures, we propose intensive fusion, which allows fusing multiple complex operators together.
Intensive fusion can further improve the inter-complex-operator data locality without the strict constraints on the subgraph structure, \textit{i.e.}, only one complex operator is allowed in prior schemes.

Unfortunately, the profits from intensive fusion are not free.
Since a complex operator involves reduction, the fusion-after-tiling optimization path often induces redundant computation.
%
%

In this subsection, we first systematically analyze the inter-operator data dependency to characterize redundant computation.
Based on the analysis, we will depict how to achieve intensive fusion efficiently by identifying two categories of subgraph structures without redundancy. 
%

\subsubsection{Why Re-computation Happens}
Suppose we have two 2-d convolution operators in a subgraph $S_i$, and now we try to fuse them.
\begin{figure}
\begin{minted}[fontsize=\footnotesize]{python}
for n, o2, h, ow in range(N, O2, H2, W2 // 16):
 # intra-tile loops, compute a tile of Conv1
 for o1, ih1, iw1 in range(O1, 1 + R2 - 1, 16 + C2 - 1):
  Conv1[n, o1, h+ih1, ow*16+iw1] = 0.0
  for ri, rr1, rc1 in range(I, R1, C1): # reduction
   Conv1[n, o1, h+ih1, ow*16+iw1] += ...
 # intra-tile loops, compute a tile of Conv2
 for io2, ih2, iw2 in range(1, 1, 16):
  Conv2[..., ow*16+iw2] = 0.0
  for ro, rr2, rc2 in range(O1, R2, C2): # reduction
   Conv2[...] += \
    Conv1[..., ow*16+iw2+rc2] * Weight2[...]
\end{minted}
\vspace{-0.05in}
\caption{Intensive fusion program of two convolutions.}
\vspace{-0.25in}
\label{code:conv_conv_fuse}
\end{figure}
In \cref{code:conv_conv_fuse}, we use $(N, O_1, H_1, W_1)$ and $(N, O_2, H_2, W_2)$ to represent the shapes of the upstream \code{Conv1} tensor and the downstream \code{Conv2} tensor, respectively. In this case, we also have $H_1 = H_2 + (R_2 - 1)$ and $W_1 = W_2 + (C_2 - 1)$ according to the convolution algorithm. For simplicity, assume the tiling of \code{Conv2} is $1\times1\times 16$ on $O_2\times H_2 \times W_2$ dimensions ($W_2 > 16$). A tile is a fraction of some tensor. Namely, the intra-tile loops \{\code{io2}, \code{ih2}, \code{iw2}\} only compute a vector of length $16$ of \code{Conv2}. According to the convolution algorithm, this vector-like tile requires a $O_1\times \left(1 + (R_2-1)\right)\times \left(16 + (C_2 - 1)\right)$ tile of \code{Conv1} for computation, which is provided by the \{\code{o1}, \code{ih1}, \code{iw1}\} loops. 
However, the reduction loops of the upstream convolution (\textit{i.e.}, the \code{ri}, \code{rr1}, \code{rc1} loops) will be executed $N \times O_2 \times H_2 \times \dfrac{W_2}{16} \times O_1 \times R_2 \times (15 + C_2)$ times in total, which is much larger than the non-fusion case $N \times O_1 \times (H_2 + R_2 - 1) \times (W_2 + C_2 - 1)$. 
%
%

We here formally depict the redundancy issue. We denote the global iteration space spanned by the loops of the upstream operator and the downstream operator as $GS_1$ and $GS_2$. Then the amount of computation can be calculated as $|GS_1|$ and $|GS_2|$ respectively. After loop tiling, we denote the iteration space spanned by the inner intra-tile loops as $TS_1$ and $TS_2$. 
Then the amount of computation is $\left|\frac{GS_1}{TS_1}\right|\cdot|TS_1|$ and $\left|\frac{GS_2}{TS_2}\right|\cdot |TS_2|$ respectively. Here we use $\frac{(\cdot)}{(\cdot)}$ as the inverse operator of Cartesian product $\times$. Next, after loop fusion, the intra-tile loops for $TS_1$ of the upstream operator will be attached to the loop structure $\frac{GS_2}{TS_2}$ of the downstream operator. Thus the iteration space size of the upstream operator can be derived as $\left|\frac{GS_2}{TS_2} \times \left(\frac{GS_1}{TS_1} - \frac{GS_2}{TS_2}\right)\right| \cdot |TS_1|$. This formula is larger than $|GS_1|$ (\textit{i.e.}, redundancy) in two cases: 1) $\frac{GS_2}{TS_2} - \frac{GS_1}{TS_1} \neq \emptyset$ (\textit{i.e.}, $\frac{GS_2}{TS_2}$ contains a loop that is not needed by $\frac{GS_1}{TS_1}$); 2) $|TS_2| < |TS_1|$, which is only determined by the data mapping function of the downstream operator.


For \cref{code:conv_conv_fuse}, the outermost iteration space $\frac{GS_2}{TS_2}$ is spanned by loops \{\code{n}, \code{o2}, \code{h}, \code{ow}\}. The middle part $\frac{GS_1}{TS_1} - \frac{GS_2}{TS_2}$ is an empty set, where $\frac{GS_1}{TS_1}$ consists of \{\code{n}, \code{h}, \code{ow}\}. And, the innermost $TS_1$ involves loops \{\code{o1}, \code{ih1}, \code{iw1}\}. The first source of redundancy is the loop \code{o2} $\in \left(\frac{GS_2}{TS_2} - \frac{GS_1}{TS_1}\right)$, thus \code{Conv1} is recomputed for $O_2$ times since a tile of \code{Conv1} is reused by $O_2$ channels/tiles of \code{Conv2}. Moreover, \code{Conv1} is recomputed for $\frac{H_2W_2\times R_2\times(15 + C_2)}{H_1W_1\times 1\times 16}$ times on the $H_2 \times W_2$ dimensions because sliding-window operations in convolution have overlaps on $H_1$ and $W_1$ of \code{Conv1} such that $|TS_2| < |TS_1|$. And an overlapping region is reused by multiple tiles of \code{Conv2}.


\begin{figure}
	\centering
	\includegraphics[width=0.45\textwidth]{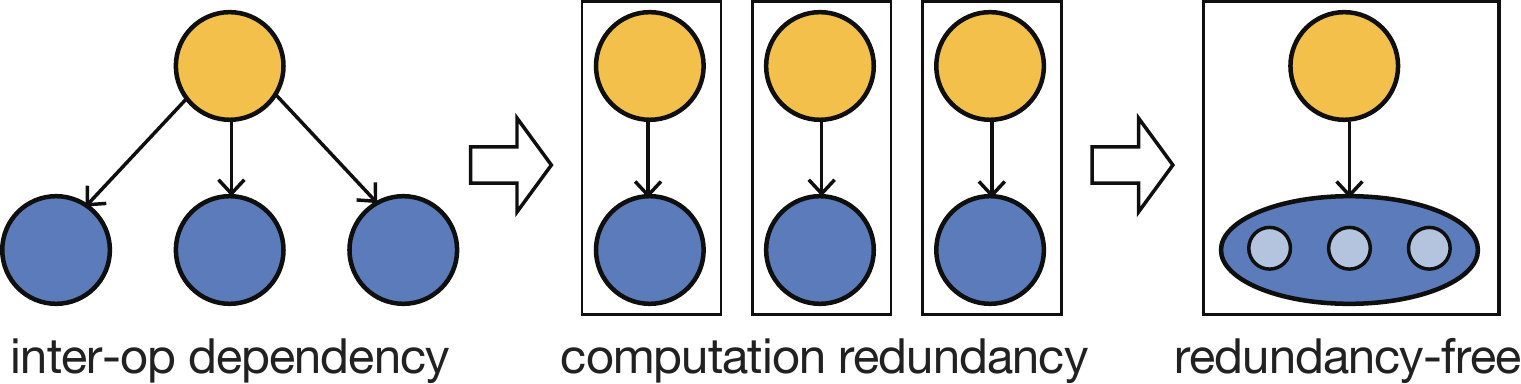}
	\vspace{-0.05in}
	\caption{Redundant computation after tiling and fusion.}
	\vspace{-0.25in}
	\label{fig:redundant_comp}
\end{figure}
We further illustrate the above issue in \cref{fig:redundant_comp}, where the yellow circle represents the output element $e$ of the upstream operator, while the blue circles denote the output elements $\{d_1, d_2, d_3\}$ of the downstream operator. 
Suppose the three blue elements reside in three separate data tiles after loop tiling. 
Then, with fusion, the yellow circle will be duplicated and stitched into three blue tiles, which leads to re-computation for three times. For the example in \cref{code:conv_conv_fuse}, $\{d_1, d_2, d_3\}$ can represent the $O_2$ output channels.

%
When $\{d_1, d_2, ..., d_k\}$ are distributed in two or more data tiles, $e$ may be re-computed for each tile, thus leading to computation redundancy.

\subsubsection{Removing the Re-computation}
As above, the re-computation will only occur under two conditions: 1) data reuse for the output tensor of the upstream operator; 2) two or more output data tiles in the downstream operator. 
The first condition cannot be removed, since it is determined only by the operator definition (\eg, convolution algorithm). 
By contrast, we can break the second condition by computing the downstream complex operator without loop tiling on the reused dimensions. 
Consequently, as the rightmost side in \cref{fig:redundant_comp} illustrates, three blue elements reside in the same tile, and the yellow circle has only one out-going edge. 

For common convolutions, the input tensor (\textit{i.e.}, the output tensor of the upstream operator) will be reused in three dimensions $O_2, H_2, W_2$ as that in \cref{code:conv_conv_fuse}. 
Without tiling, the original output data size of $O_2\times H_2 \times W_2$ will normally be larger than the cache capacity, hence poor cache utilization during execution.
Fortunately, there are widely used convolution operators on mobile devices without this concern, namely, depthwise convolution and pointwise convolution. 
The former does not perform reduction on the input channel dimension while the latter is free of reduction in kernels (\textit{i.e.}, $R_2=C_2=1$). 
Specifically, the input tensor will be reused only on $H_2, W_2$ dimensions in depthwise convolution, and only $O_2$ dimension in pointwise convolution.

\begin{figure}
    \centering
	\subfigure[Downstream depthwise conv.]{\label{subfig:dep_tiling}{\includegraphics[width=0.21\textwidth]{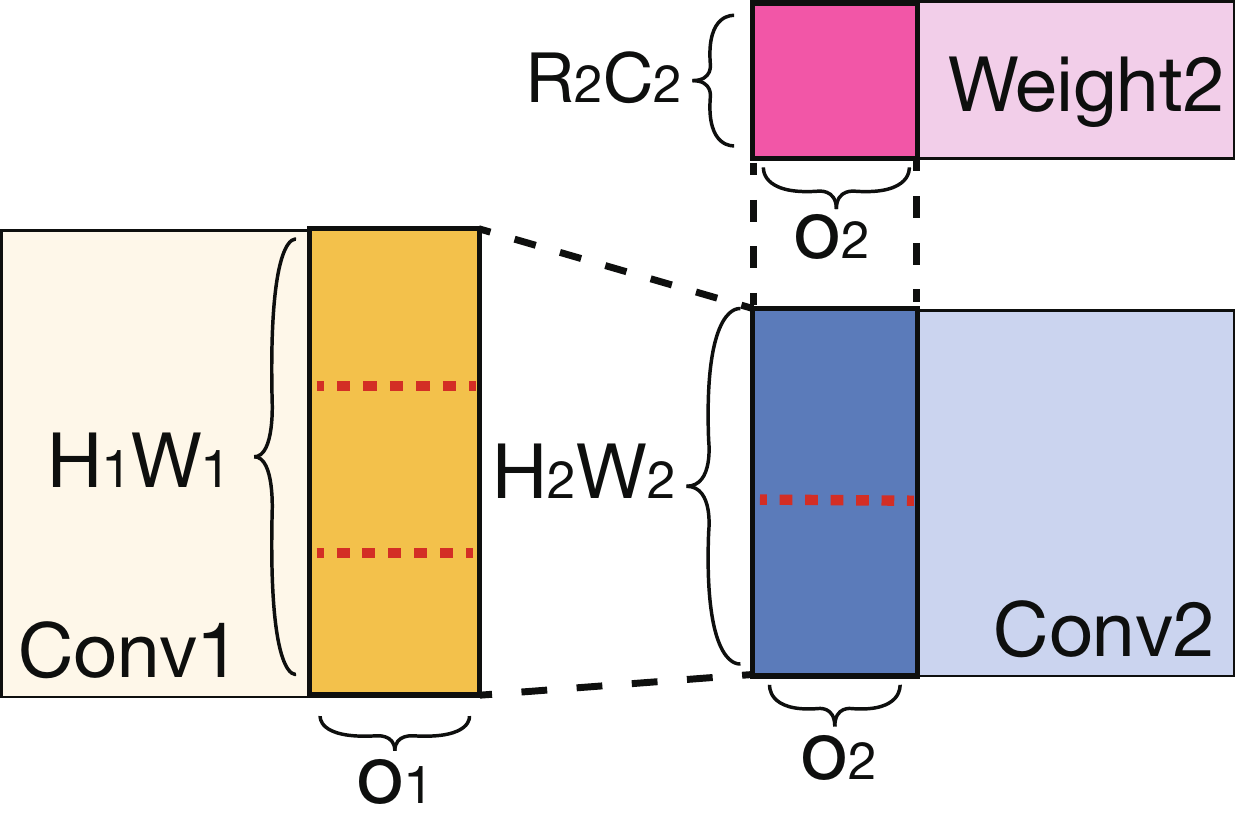}}} 
    ~
	\subfigure[Downstream pointwise conv.]{\label{subfig:1x1_tiling}{\includegraphics[width=0.21\textwidth]{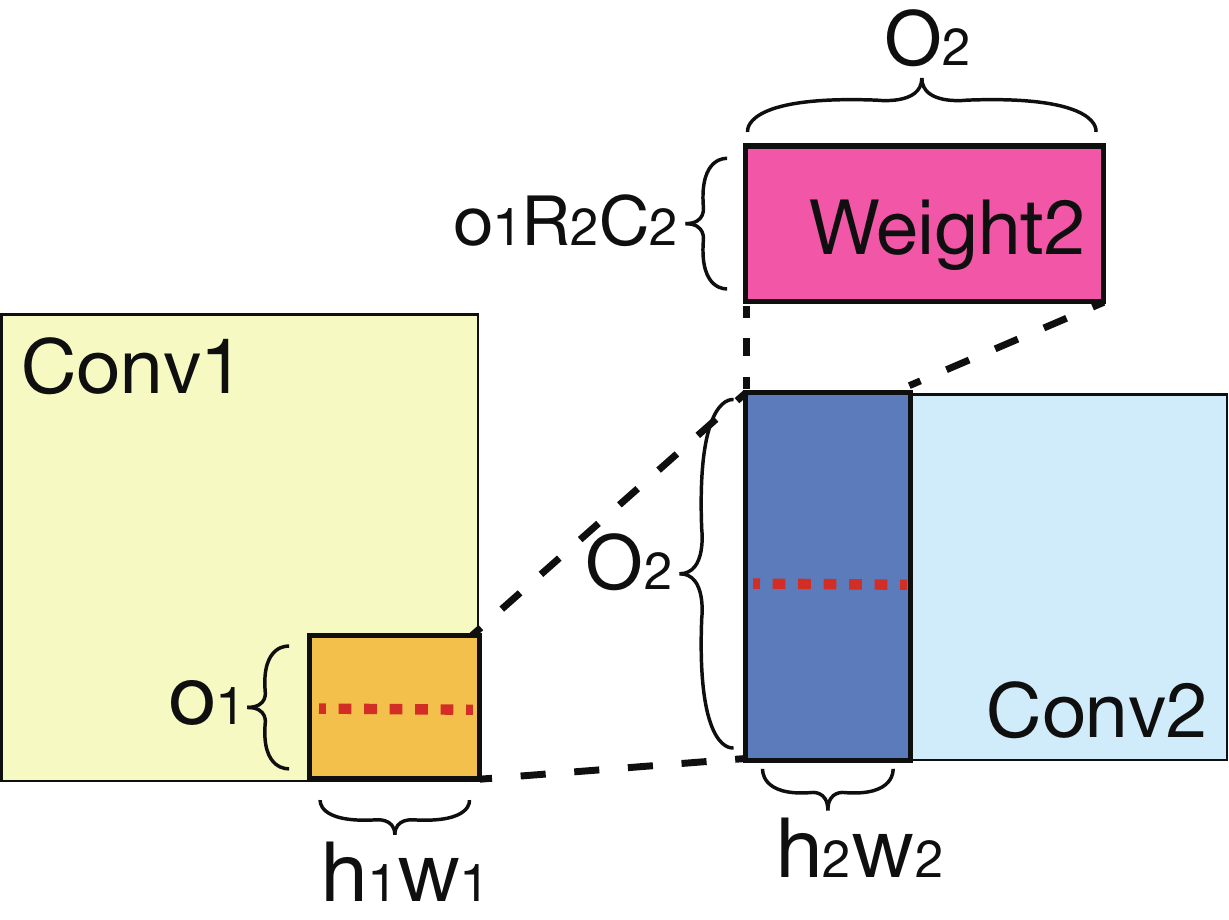}}}%
	\vspace{-0.05in}
	\caption{Inter-operator data dependency in intensive fusion.}
	\vspace{-0.2in}
	\label{fig:deeper_fusion_tiling}
\end{figure}

We achieve intensive fusion for these two categories in \cref{fig:deeper_fusion_tiling}. We use uppercase letters to denote the original dimensions, while the corresponding lowercase letters to denote the tiled dimensions.
When the downstream convolution is depthwise, as shown in \cref{subfig:dep_tiling}, its data tile should be $H_2W_2o_2$, where $H_2, W_2$ dimensions are not tiled due to reuse and $o_2$ is a tiled dimension from $O_2$. 
Then the data tile of the upstream convolution should be $H_1W_1o_1$. In this case, we will also have $o_1=o_2$ since the number of input channels is the same as the number of output channels in the downstream depthwise convolution. Similarly, the tile of the downstream pointwise convolution can be denoted as $h_2w_2O_2$ in \cref{subfig:1x1_tiling}. 
Both of them is much smaller than $O_2\times H_2 \times W_2$. Notably in \cref{fig:deeper_fusion_tiling}, the inter-operator data mapping is determined by the convolution algorithm. For example, in \cref{subfig:1x1_tiling}, the \code{Conv2} tile $h_2w_2O_2$ requires a \code{Conv1} tile $h_1w_1o_1$ and a \code{Weight2} tile $o_1R_2C_2O_2$ for computation, where $h_1=h_2$, $w_1=w_2$, and $R_2=C_2=1$. 
%
Additionally, no constraints are imposed on the inner-level tiling of the two convolutions, indicated as the red dashes in \cref{fig:deeper_fusion_tiling}. 
In other words, if intensive fusion requires that two convolution tiles reside in L1 cache for locality, then the tiling of themselves on registers will not be affected. 
Putting all the analysis together, our intensive fusion creates new optimization opportunities for complicated subgraphs, when the downstream convolution is depthwise or pointwise. 
We do not include the analysis of matrix multiplication because it is mathematically equivalent to pointwise convolution.
Even if the downstream convolution type is unmet for intensive fusion, our tuner can still benefit from a larger tuning space given a complicated subgraph. 
Meanwhile, the tuning efficiency for such unmet complicated subgraphs will be addressed by the later reformer layer.
Therefore, our tuner eschews the need of prior constraints on subgraph structures in favor of intensive fusion and joint optimization.

\postsec
\presec \section{Graph Partitioning in Frontend}\label{sec:part}\postsec
Given the original directed computational graph $G$, the graph frontend will partition $G$ into many smaller subgraphs, each of which can be optimized separately and efficiently.
Here, we refer to the graph partition as a collective term.
It is formally defined as a set of subgraphs $S_1, S_2, ..., S_n$, such that the nodes in each subgraph are disjointed and each node in $G$ belongs to exactly one subgraph.
Based on our powerful tuner, we allow an arbitrary structure for each $S_i$.
This further indicates that any edge in $G$ can potentially cross the cut in the partition.
However, this can lead to 1) search space explosion and 2) cycles in the resulting graph partition.
Larger search space will increase the optimization complexity, thus decrease the tuning efficiency for the tuner backend.
To strike this issue, our solution involves two aspects.
First, in the graph frontend, our partitioning algorithm will assign a weight for each operator and control the weight of each subgraph, hence obviating unreasonably huge subgraphs.
Second, in the later reformer layer, we design a divide-and-conquer tuning mechanism to reduce the tuning complexity for each subgraph. 
Another issue is the cyclic dependency among subgraphs, which may disable some critical optimization techniques, \textit{e.g.}, data layout selection, and lead to deadlocks during runtime execution.
To generate a cycle-free partition, we devise an iterative clustering algorithm, which will theoretically guarantee the acyclic property of the graph partitioning.

\presub \subsection{Weight Assignment for Operators} \label{sec:pig}\postsub
\begin{figure}
	\centering
	\includegraphics[width=0.34\textwidth]{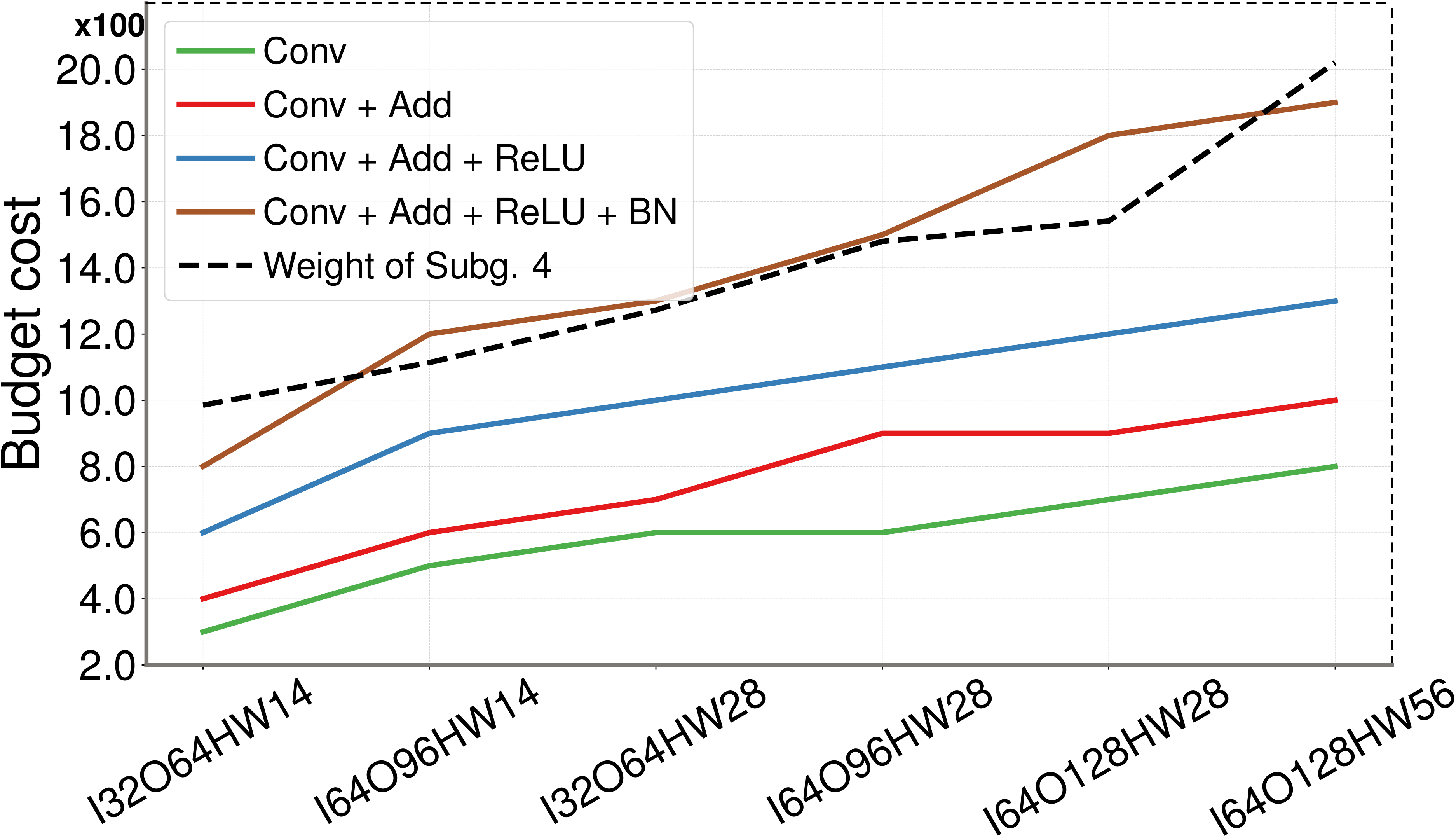}
	\vspace{-0.1in}
	\caption{Budget costs of optimizing different subgraphs.}
	\vspace{-0.2in}
	\label{fig:weight_cost}
\end{figure}

In a computational graph $G$, we denote the set of nodes as $V$ and the set of directed edges as $E$. 
Each node in the graph, denoted as $v$, represents an operator in the model, and each of the directed edges, denoted as $e$, represents a tensor produced by the source operator and consumed by the destination operator. 
%
%


%

To avoid an unreasonably huge subgraph $S_i$, we resort to measuring the tuning complexity of $S_i$ directly during partitioning. 
Previous works use indirect metrics as weights \cite{roesch2018relay}, \textit{e.g.}, the number of operators in a subgraph $S_i$, which we find ineffective.
Specifically, we observe that the contributions of operators in $S_i$ towards the tuning complexity are not the same and highly depend on their tensor shapes and operator types.
We here conduct an experiment to study the relationship between the subgraph structure and the tuning complexity. 
We use \emph{tuning budget}, which is the total number of explored schedules to obtain stable performance for a subgraph, as an indicator of the tuning complexity.
We then tune different subgraphs, each of which contains different operators, and record their tuning budgets.

We report the results in \cref{fig:weight_cost}, where the budget is on a scale of 100, the batch size is 1, the padding size in convolution is 1, the height/width of the convolutional window is 3, and the numbers behind $IOHW$ are the sizes of other corresponding dimensions. For example, in the second subgraph ($Conv + Add$), the input tensor has shape ($N$=1, $I$=32, $H$=28, $W$=28), the output tensor of $Conv$ operator has shape ($N$=1, $O$=64, $H$=28, $W$=28).
%
Based on \cref{fig:weight_cost}, we have two observations. 
First, for each subgraph structure, the tuning budget does not scale directly with the number of operators, but illustrates a linear trend with tensor shapes.
Second, for a given tensor shape, the tuning budget scales almost linearly with the number of operators, although the tuning space size increases exponentially with the number of operators.

The root cause of the first observation is that the most major optimization technique during tuning is loop transformation, \eg, loop tiling and fusion in \cref{sec:deeper}. 
%
%
Thus the tuning complexity is directly determined by the loop nest in the program for the operator. 
This further involves two folds: 1) the number of loops (\textit{e.g.}, seven nested loops in a 2-d convolution); 2) the extent of each loop.
Thus, we define a fine-grained weight for each operator as follows, which measures the tuning complexity as a linear function of the loop nest: 
%
\begin{equation}
w_v= c \times \prod_{l \in L_v}{\log(s_l)} + b\, ,
\label{eq:weight}
\end{equation}
\vspace{-0.03in}
where $L_v$ is the set of loops for the operator $v$, $s_l$ is the extent of the loop $l \in L_v$, while $c$ is the slope and $b$ is the bias. With \eqref{eq:weight}, it is easy to see a larger weight indicates higher complexity of tuning. 
Then, according to the second observation, the weight of a subgraph $S_i$ can be derived as the sum of the weights of all operators in $S_i$. 
As illustrated by the black dash line in \cref{fig:weight_cost}, we can almost perfectly fit the tuning budget with \cref{eq:weight}.
Subsequently, we are able to guarantee a tractable size for each subgraph by setting up a threshold as the maximum weight. 
Moreover, such design helps eliminating trivial subgraphs that may waste tuning budgets and yielding balance among all subgraphs.

\vspace{0.01in}
\presub\subsection{Acyclic Partitioning}\postsub

\begin{figure}
	\centering
	\includegraphics[width=0.28\textwidth]{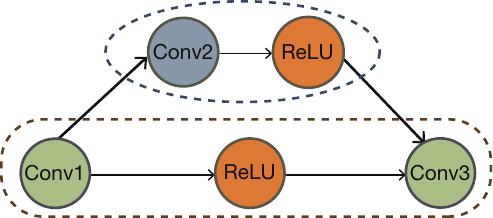}
	\vspace{-0.1in}
	\caption{Cyclic dependency with complicated subgraphs.}
	\vspace{-0.25in}
	\label{fig:cyclic_partition}
\end{figure}

After calculating weights, we here propose a new algorithm to address the issue of cyclic dependency.
To allow arbitrary subgraph structures, the graph frontend can incur cycles in the resulting partition unexpectedly.
For example, suppose $Conv1$ and $Conv3$ in \cref{fig:cyclic_partition} can trigger the intensive fusion.
Then we put them in the same subgraph $S_1$, while $Conv2$ constitutes another subgraph $S_2$. 
%
%
In this case, $S_1$ and $S_2$ have inter-dependency in input tensors and output tensors.
%
%
Such cycles can lead to deadlocks when executing these subgraphs at the runtime.
Although prior works employ heuristics to produce subgraphs without cycles \cite{roesch2018relay, apollo}, the generated subgraphs are over simplified and many opportunities are thus excluded. 
For example, three complex operators in \cref{fig:cyclic_partition} will be placed into three separate subgraphs \cite{roesch2018relay, apollo}, hence missing opportunities of intensive fusion and joint optimization.


%

We first formally define the acyclic property for a graph partition. 
%
%
We call a partition without cycles a \emph{$n$-way acyclic partition}, if it satisfies the following property.
%
\vspace{-0.04in}
\begin{Definition}\label{def:kacyc}\emph{$n$-way acyclic partition}: A $n$-way acyclic partition contains $n$ disjoint sets of nodes $\{V_1,V_2,...V_n\}$.
For any $u,u' \in V_i$, $v,v' \in V_j$, $1 \leq i \neq j \leq n$, there cannot exist two paths, from $u$ to $v$ and from $v'$ to $u'$, at the same time.
\end{Definition}
\vspace{-0.04in}
%
Further, we introduce the concept of \textit{topological stage} as an identifier to the topological position of each node in $G$.
\vspace{-0.04in}
\begin{Definition}\label{def:topstage}\emph{topological stage}: The topological stage $ts_v \geq 1$ is an integer denoting the position of $v$ in $G$. It can be calculated as the length of the \emph{longest path} from the root $r$ (a node with zero in-degree) to the current $v$.
\end{Definition}
\vspace{-0.04in}
It is easy to see, for any node $v \in V$, $\forall e = (u, v)\in E$, we have $ts_u < ts_v$; $\forall e = (v, w) \in E$, we have $ts_w> ts_v$.

Based on the concept of topological stage, we observe that for each node $v$, there exists a set of nodes that can be safely grouped with it. 
We call such a special set \emph{affix set}.
\vspace{-0.04in}
\begin{Definition}\label{def:tas}
\emph{affix set}: 
%
We first denote the set of nodes that $v$ can connect to in the underlying undirected graph corresponding to $G$ as $UC_v$. 
Then, the affix set $AS_v$ for node $v$ is a subset of $UC_v$, such that each node in ${AS}_v$ satisfies one of the following two conditions:
\vspace{-0.06in}
\begin{eqnarray}
    &\forall{u} \in {AS}_v,&\,{ts}_u = {ts}_v+1;\nonumber\\
    &\forall{u} \in {AS}_v,&\,{ts}_u = {ts}_v-1.\nonumber
    \vspace{-0.1in}
\end{eqnarray}
\end{Definition}
\vspace{-0.04in}
With the above definitions, we can derive the following theorem, which is the core of our later graph partitioning algorithm to guarantee the acyclic property.
\vspace{-0.03in}
\begin{theorem}\label{thm:acyc} Given a node $v$ and its affix set ${AS}_v$, there will exist no cycles in the resulting graph partition if $v$ and any nodes in ${AS}_v$ cluster together to produce a new subgraph.
\end{theorem}
\vspace{-0.1in}
\begin{proof} 
We will prove the theorem by deducing a contradiction.
Specifically, for any two nodes $u$ and $v$ in $G$, assume that grouping $u$ and $v$ together will produce a cycle in the partition. 
Since the original graph $G$ is acyclic, the generated cycle indicates that there must exist a path $u \rightarrow p \rightarrow v$ in $G$, where $p$ is another node.
For example, in \cref{fig:cyclic_partition}, the $Conv1, Conv2, Conv3$ operators are $u, p, v$, respectively.
%
%
However, we only group $u$ and $v$ if $u \in {AS}_v$ or $v\in {AS}_u$. 
Thus, we have $|ts_v - ts_u|=1$, which means the path from $u$ to $v$ or $v$ to $u$ must have no other nodes, \textit{i.e.}, $p$ does not exist.
In summary, no cycle will be generated if $v$ and a node $u \in {AS}_v$ cluster together to generate a new subgraph.
\end{proof}

\begin{algorithm}[t]
    \caption{Graph Partition} 
    \label{alg}
    \begin{algorithmic}[1]
        \Function{Cluster}{$G$} \label{clusters}
            \State Initialize $Td, Cand, TopStage$ \label{init}
            \State Calculate weight for each $v$ in $G$ \label{assignw}
            \While{$Cand \neq \emptyset$} \label{startgreedy}
              \State Choose $v \in Cand$ with heaviest weight \label{selectv}
              \If{$\exists u \in AS_v, \textit{s.t.}, w_v + w_u < Td$} \label{clusterstart}
                \State $u$ and $v$ cluster together as a hyper node $v'$ \label{ln:gen_subg}
                \State Move $v'$ to $Cand$ \label{updatecande}
              \Else
                \State Remove $v$ from $Cand$ \label{skip}
              \EndIf \label{clusterend}
              \State Update $E, TopStage$ \label{update}
            \EndWhile \label{endgreedy}
            \State Translate hyper nodes into subgraphs $\{S_1,S_2,...\}$
            \State \textbf{return} $\{S_1,S_2,...\}$
        \EndFunction \label{clustere}
    \end{algorithmic}
\end{algorithm}
\vspace{-0.03in}

Based on \cref{thm:acyc}, we can derive our cycle-free graph partitioning algorithm, in which affix operators iteratively cluster together to yield subgraphs.
We illustrate the \textsc{Cluster} algorithm in \cref{alg}. 
%
At \cref{init}, we pre-process the graph $G$ to initialize some data structures, including the maximum weight threshold $Td$ for subgraphs, the initial candidate node set $Cand$ that contains all nodes in $G$, and the information on topological stages $TopStage$. Then we calculate the weight for each operator at \cref{assignw}.
%
%
Next, we group affix nodes to generate subgraphs, and the weight of each subgraph is controlled via a greedy strategy (\cref{startgreedy} - \cref{endgreedy}).
In each iteration, a node $v$ (a subgraph $S_i$ can be viewed as a hyper node $v$) with the heaviest weight in $Cand$ is selected (\cref{selectv}). 
%
We then search in $AS_v$ to find a node $u$ with the smallest weight.
If the sum of weights of $v$ and $u$ is smaller than the threshold $Td$, a new subgraph will be produced.
This subgraph is also viewed as a new hyper node $v^\prime$, and put in the candidate set for further clustering (\cref{ln:gen_subg} - \cref{updatecande}).
Otherwise, the node $v$ will be skipped and removed from $Cand$ (\cref{skip}).
%
%
After each iteration, we will update the edge set $E$ and the position information $TopStage$ (\cref{update}).
The algorithm will repeat until the candidate set is empty.
In this way, the resulting partition is guaranteed to be acyclic and each subgraph has a reasonable weight that is smaller than $Td$.

%


\vspace{-0.03in}
\presec \section{Divide-and-Conquer Tuning in Reformer Layer} \label{sec:rpadc}\postsec

The search space of a subgraph with a complicated structure is still large even if we limit its weight during graph partitioning. 
To achieve efficient tuning, we insert a reformer layer between the graph frontend and the tuner backend to orchestrate different components. 
The reformer layer exploits a divide-and-conquer mechanism to break down the complicated subgraph tuning into smaller sub-tasks and then combine them. 
%
As the dividing stage, we design a \textsc{Split} function. 
By invoking \textsc{Cluster} in \cref{alg}, the \textsc{Split} function further splits each large subgraph $S_i$ into several mini-subgraphs $\{M_{i1}, M_{i2}, ...\}$. Each mini-subgraph has at most one complex operator and a smaller weight.
%
Then, as the conquering stage, we devise a \textsc{Join} function.
It can combine those mini-subgraphs $\{M_{i1}, M_{i2}, ...\}$ back to $S_i$ for further tuning.
%

%
%


The reformer layer will immediately execute the \textsc{Split} function after graph partitioning.
Then, by inspecting the feedbacks from the tuner backend, the reformer layer will call the \textsc{Join} function to combine those mini-subgraphs $M_{ij}$ back as $S_i$, if the tuning for each $M_{ij} (1 \leq j \leq m_i)$ tends to stabilize. 
%
%
During joining, the schedules searched by the tuner for each mini-subgraph will also be composed as a large schedule for $S_i$.
When delivering $S_i$ to the tuner backend, this combined schedule will be treated as the initial schedule to evade inefficient tuning from the scratch for $S_i$.

Our cycle-free partitioning function \textsc{Cluster} only limits the maximum weight for each subgraph without other constraints on structures. Further, with the \textsc{Split} and \textsc{Join} functions, \sys is able to optimize any subgraph efficiently.
%


\presec \section{Evaluation}\label{sec:eval} \postsec

We implemented the graph frontend in C++ based on TVM (0.8dev1) \cite{chen2018tvm}, and the reformer layer and the tuner in both Python and C++.
In this section, we evaluate \sys on two mobile CPU platforms with Kirin 990 SoC (Android v10), representing high-end devices, and Qualcomm snapdragon (Qsd) 810 SoC (Android v8), representing low-end devices with strict resource constraints.
We compare \sysname with Torch Mobile \cite{paszke2019pytorch} and Ansor \cite{zheng2020ansor}. 
Torch Mobile is a widely used deep learning framework, which employs a hand-tuned high-performance library XNNPACK \cite{xnnpack} developed by Google.
While Ansor is the state-of-the-art auto-tuning framework based on TVM, which performs better than TensorFlow Lite \cite{abadi2016tensorflow, zheng2020ansor}.
Thus, TensorFlow Lite is not included in our benchmarks.
%
%
%
\vspace{0.01in}
\presub\subsection{End-to-End Performance} \postsub
\label{subsec:e2e}

\begin{figure*}[!ht]
	\centering
    \subfigure[Input tensor shape (1, 3, 56, 56).]{
        \label{sfig:sumsung 56}
    	\includegraphics[width=0.3\textwidth]{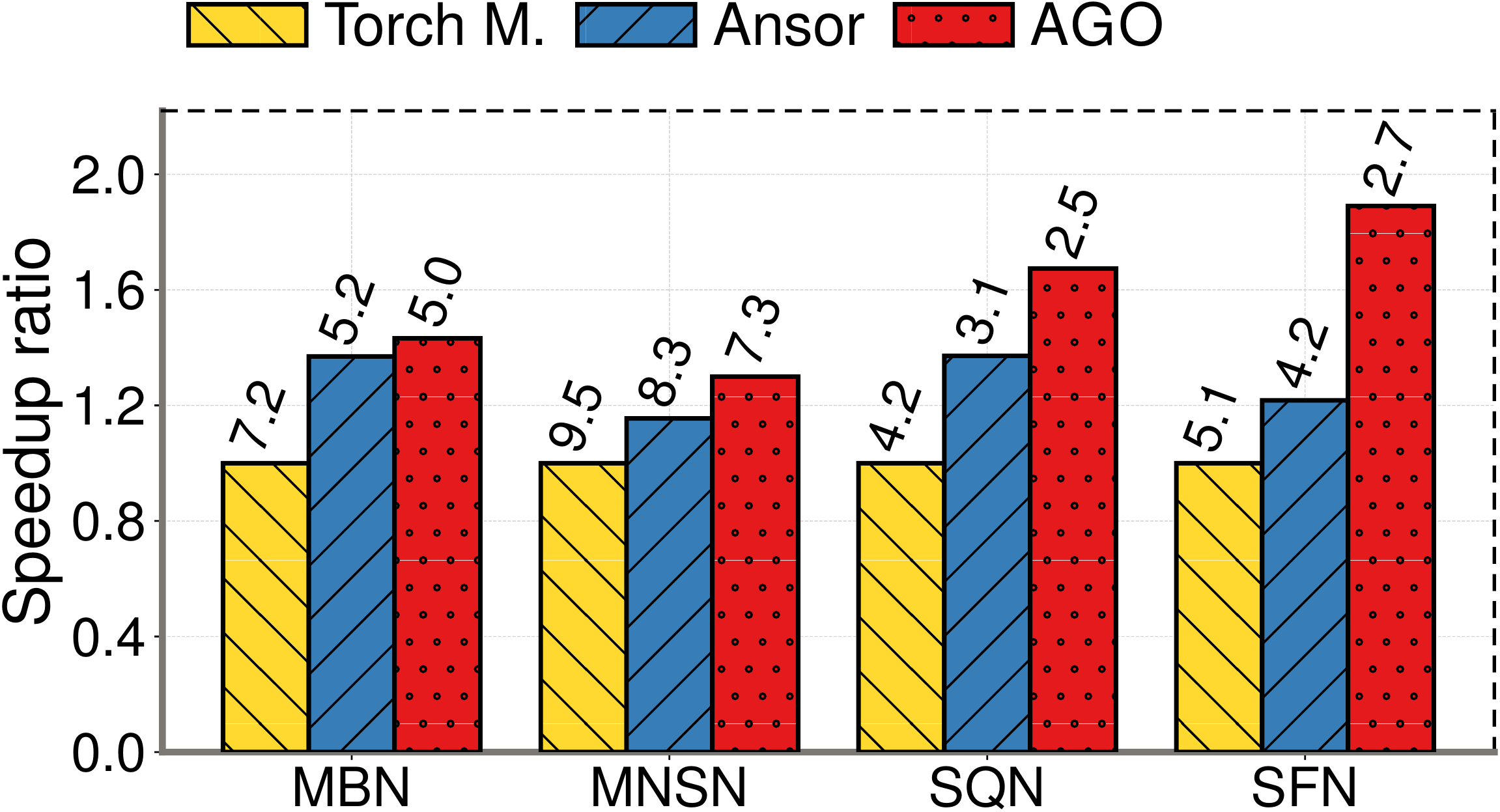}}
	\subfigure[Input tensor shape (1, 3, 112, 112).]{
		\includegraphics[width=0.3\textwidth]{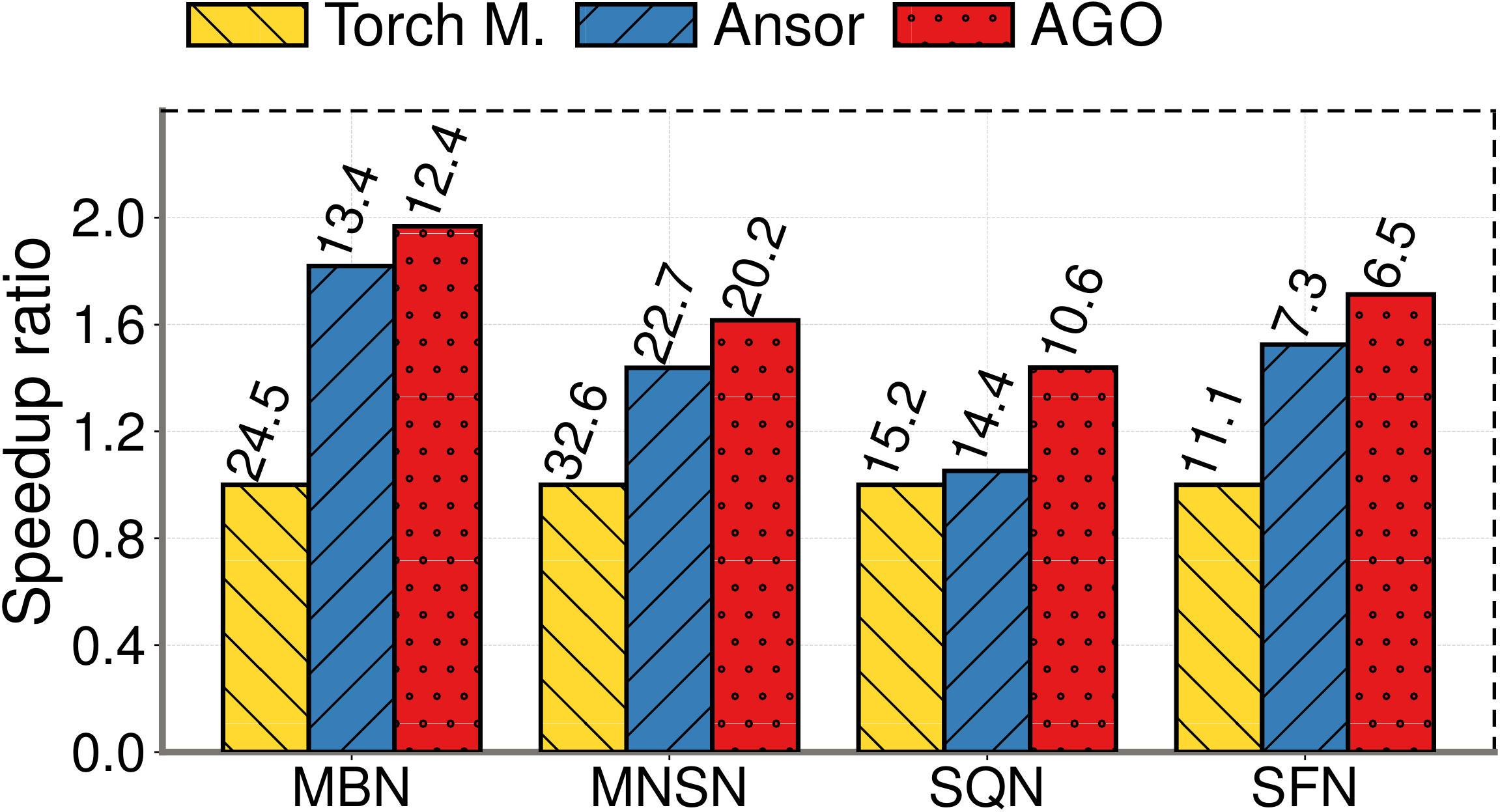}
    	\label{sfig:sumsung112}}
    \subfigure[Input tensor shape (1, 3, 224, 224).]{
    	\includegraphics[width=0.3\textwidth]{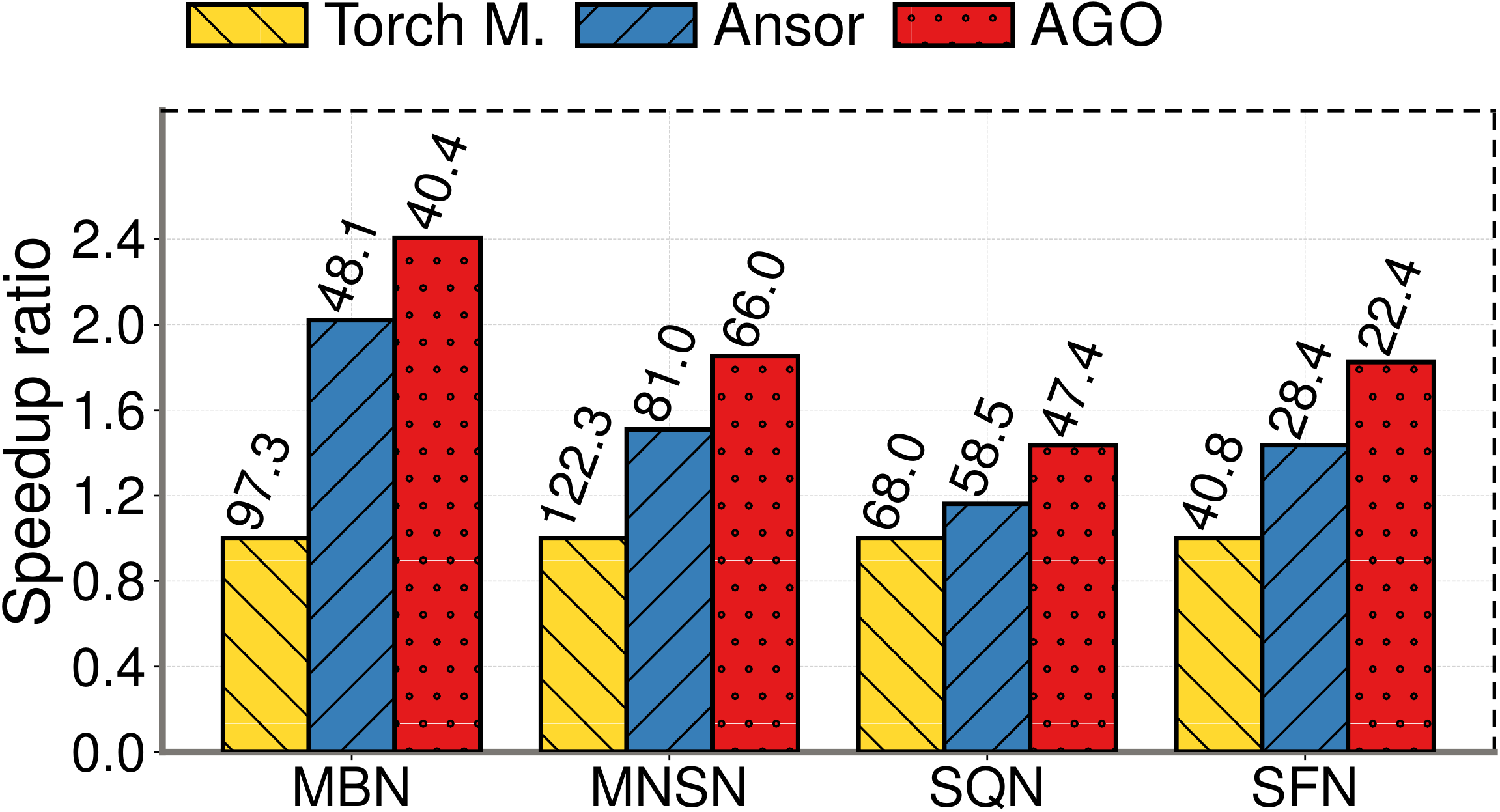}
    	\label{sfig:sumsung224}
    }
    \vspace{-0.1in}
    \caption{End-to-end inference performance on Qsd 810 SoC.}
    \label{fig:samsungbenchmark}
    \vspace{-0.2in}
\end{figure*}

\begin{figure*}[!ht]
	\centering
    \subfigure[Input tensor shape (1, 3, 56, 56).]{
        \label{sfig:huawei 56}
    	\includegraphics[width=0.3\textwidth]{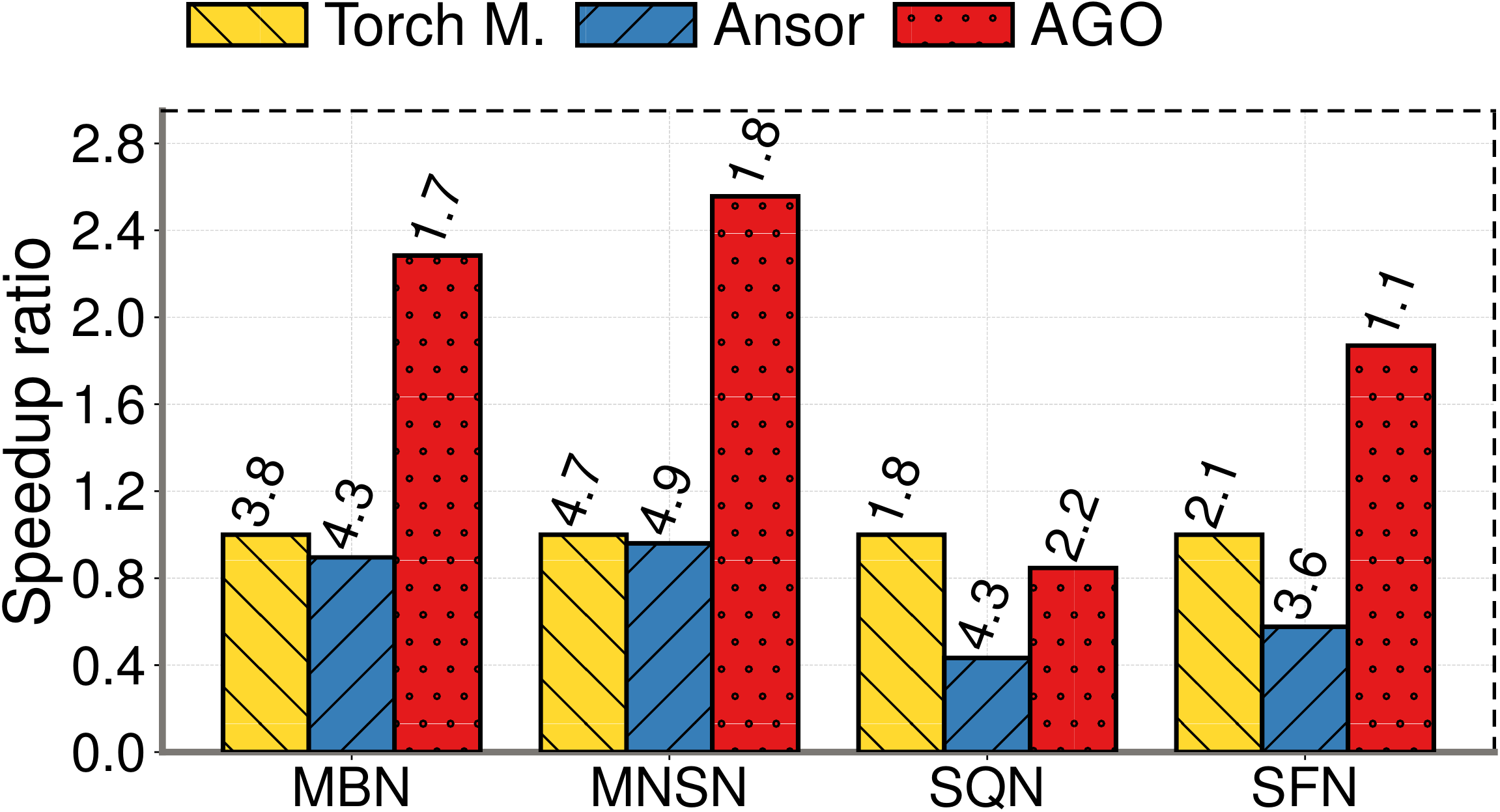}}
	\subfigure[Input tensor shape (1, 3, 112, 112).]{
		\includegraphics[width=0.3\textwidth]{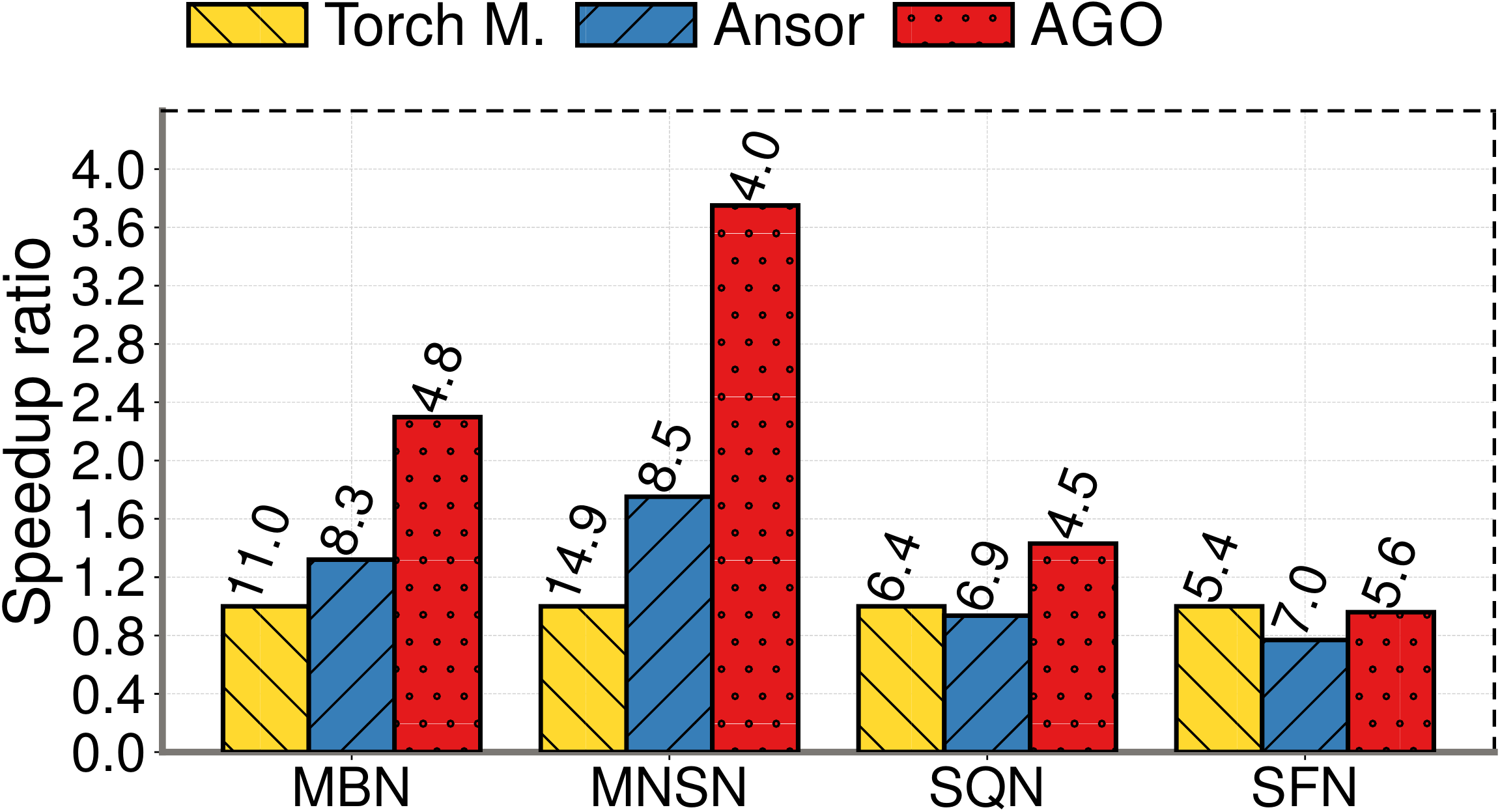}
    	\label{sfig:huawei112}}
    \subfigure[Input tensor shape (1, 3, 224, 224).]{
    	\includegraphics[width=0.3\textwidth]{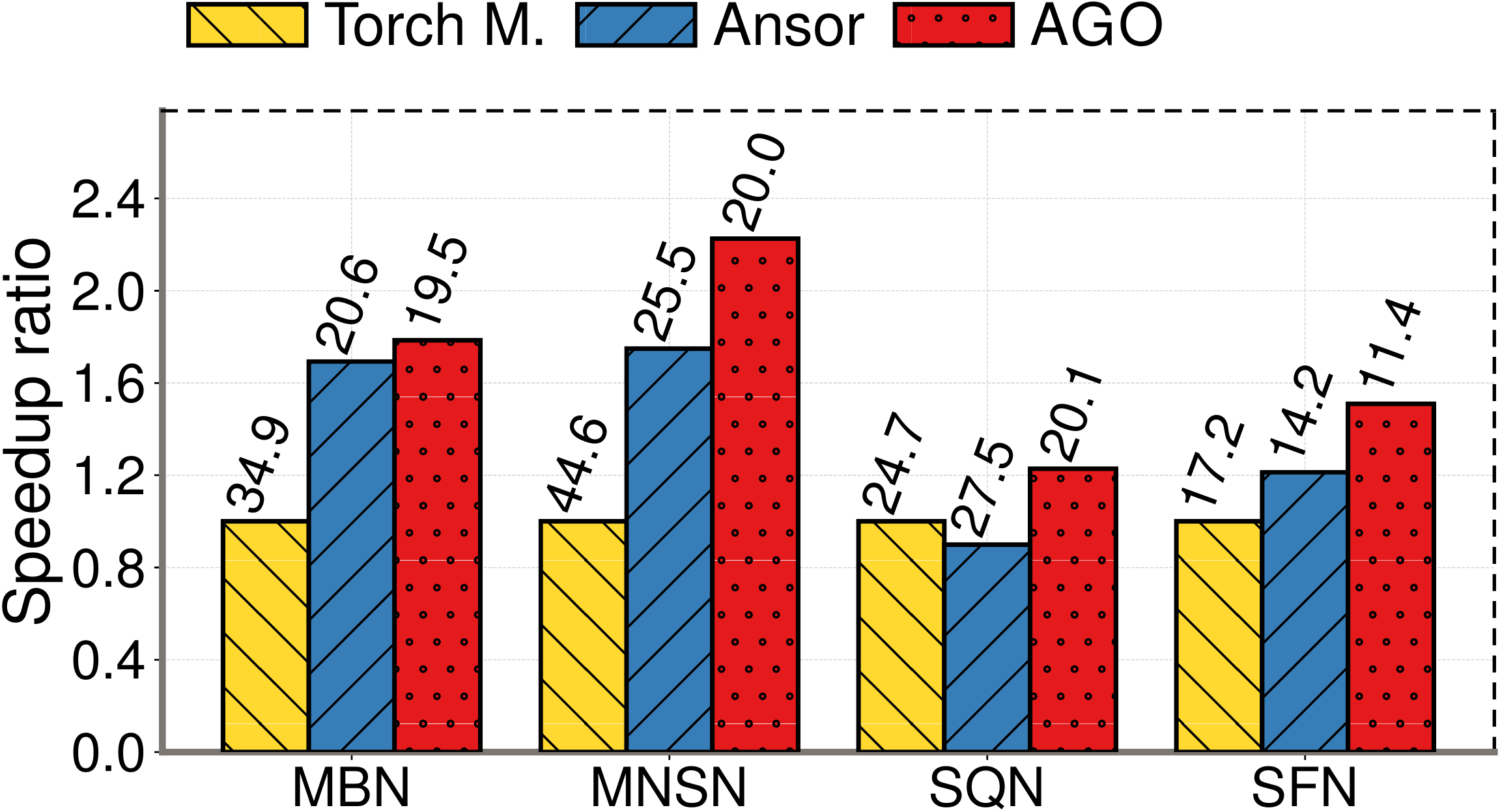}
    	\label{sfig:huawei224}
    }
    \vspace{-0.1in}
    \caption{End-to-end inference performance on Kirin 990 SoC.}
    \label{fig:huaweibenchmark}
    \vspace{-0.2in}
\end{figure*}

\begin{figure}
    \centering
    \includegraphics[width=0.39\textwidth]{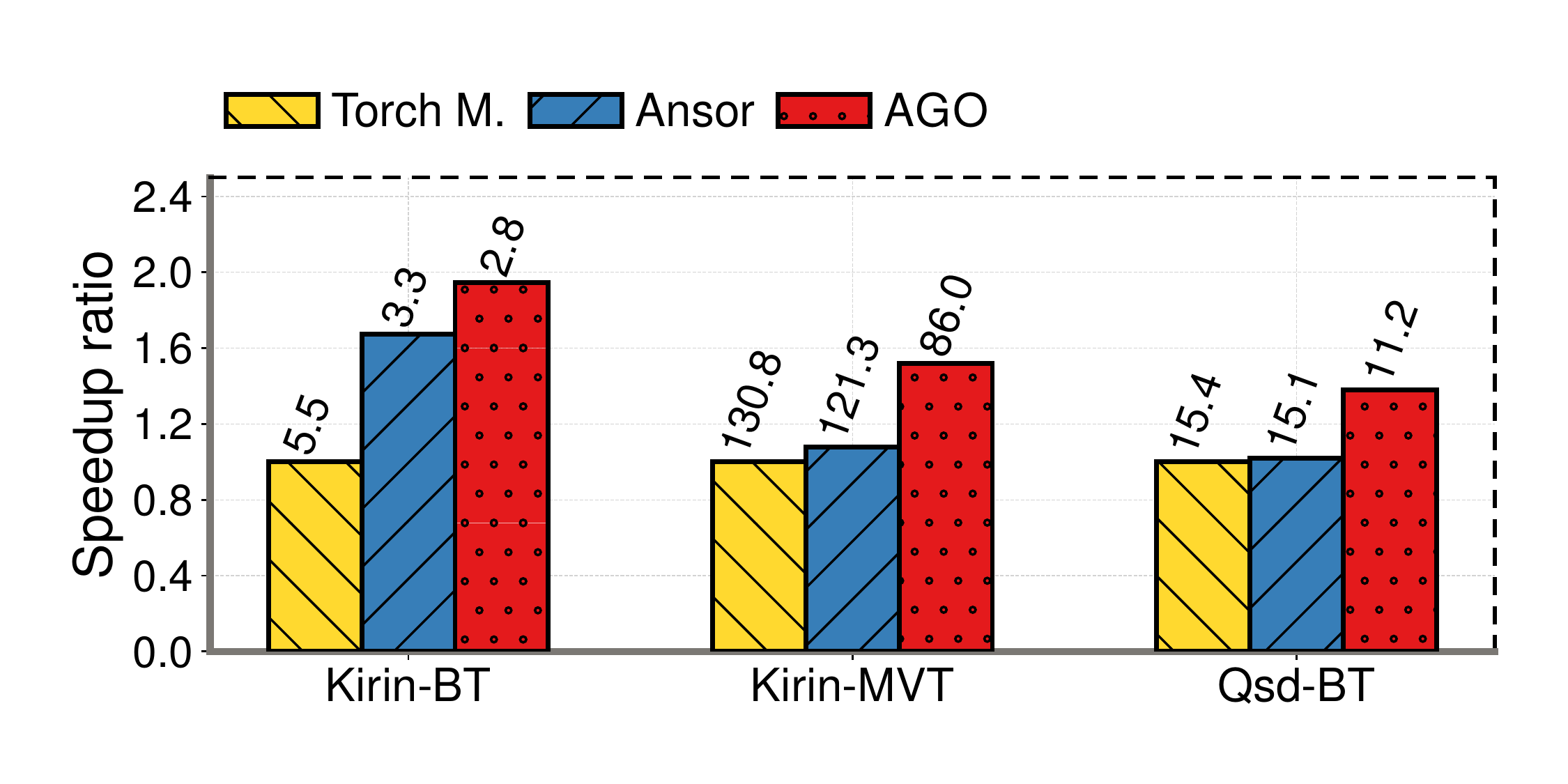}
    \vspace{-0.23in}
    \caption{End-to-end inference performance for BT and MVT.}
    \label{fig:bt_mvt_e2e}
    \vspace{-0.23in}
\end{figure}
In this subsection, we evaluate \sys in terms of end-to-end inference performance. Our benchmarks cover four classical neural networks: MobileNet-V2 (MBN) \cite{mobilenet}, MNasNet (MNSN) \cite{mnasnet}, SqueezeNet (SQN) \cite{sqeezenet}, and ShuffleNet-V2 (SFN) \cite{ma2018shufflenet}, and two emerging new networks: Bert-tiny (BT) \cite{DBLP:journals/corr/abs-1908-08962, bhargava2021generalization} and MobileViT (MVT) \cite{mehta2022mobilevit}.
These networks are lightweight and widely used for mobile deep learning services.
For both Kirin 990 SoC and Qsd 810 SoC, we set the batch size to 1 for all input tensors due to constrained computing power, which is also a general setting for mobile inference. 
Besides, we test different shapes of the input tensor for each classical network: small shape ($N$=1, $I$=3, $H$=56, $W$=56), middle shape ($N$=1, $I$=3, $H$=112, $W$=112), and large shape ($N$=1, $I$=3, $H$=224, $W$=224).
These shapes represent various workloads due to divergent image resolutions in real applications.
For the new language model BT, we set the input sequence length to 128, which is the longest sequence it supports.
For the new model MVT, we only evaluate it on the large shape, which is the image size of the Imagenet dataset \cite{deng2009imagenet}.
All networks are executed with float32 precision.
Besides, we set the search budget for \sys and Ansor to 20,000, which is suggested by Ansor \cite{zheng2020ansor} for sufficient tuning.

We report the speedup of each method over Torch Mobile for classical networks in \cref{fig:samsungbenchmark} and \cref{fig:huaweibenchmark}, where the number on the top of each bar is the raw latency in milliseconds.
On the Qsd 810 SoC, \sysname achieves an average speedup of $1.5\times$, $1.6\times$, and $1.8\times$ over Torch Mobile on three input tensor shapes respectively.
%
Compared with Ansor, \sysname achieves an average speedup of $1.2\times$ on each input shape. 
%
The main reason behind the significant improvement over Torch Mobile is that, hand-tuned libraries often put tremendous engineering efforts on optimizing typical workloads, while other non-typical operators are less optimized.
The speedup over Ansor mainly originates from the intensive fusion and joint optimization for complicated subgraphs, which are the opportunities missed by Ansor.
For example, when there are many subgraphs with consecutive pointwise and depthwise convolutions, \sysname achieves an average of $1.3\times$ speedup over Ansor.
%

Similarly, on the Kirin 990 SoC, \sysname achieves average $1.9\times$, $2.1\times$, and $1.5\times$ speedup over Torch Mobile on three input tensor shapes respectively. 
%
%
Compared with Ansor, \sysname achieves average $2.6\times$, $1.6\times$, and $1.1\times$ speedup respectively.
Again, such improvements are directly owing to our intensive fusion and joint optimization. 
By contrast, Ansor suffers from a limited tuning space due to the simple subgraph structures generated by Relay \cite{roesch2018relay}.
For example, \sys outperforms both baselines on MNSN significantly, which involves massive pointwise and depthwise convolutions. %
Both Torch Mobile and Ansor can only perform conventional fusion, while \sys can achieve either intensive fusion or joint optimization.

We further employ \sys to optimize BT and MVT, which are two new networks. We report the results in \cref{fig:bt_mvt_e2e}, where we do not test MVT on the Qsd 810 SoC due to its limited resources. Compared with Torch Mobile, \sys improves the performance by $38.2\%$ on BT and $34.3\%$ on MVT, respectively. Further, compared with Ansor, \sys improves the performance by $20.5\%$ on BT and $29.1\%$ on MVT. In summary, \sys can be used to boost new neural architectures readily without any interference.

Additionally, the tuning budget of 20,000 implies up to a day of the compilation time. But this is affordable to practitioners since they only need to execute \sys once before the long-run deployment. Moreover, it is much shorter than weeks or even months of hand-tuning.

\presub\subsection{Micro Benchmark}\postsub

\begin{figure*}
	\centering
	\subfigure[Two depthwise convolutions.]{\label{subfig:dep33_dep33_fuse}{\includegraphics[width=0.24\textwidth]{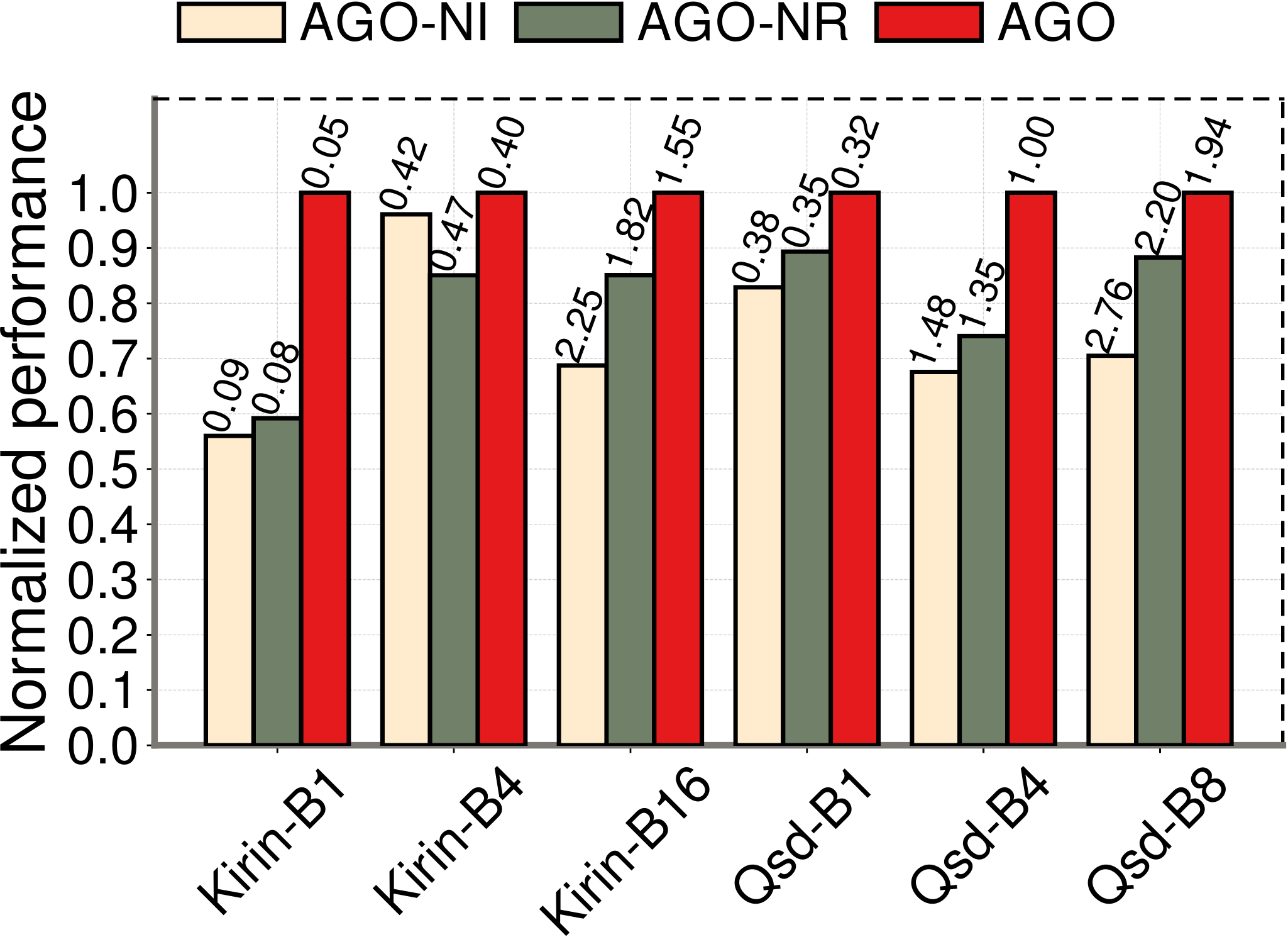}}}\hfill
	~
	\subfigure[Depthwise conv \& pointwise conv.]{\label{subfig:dep33_conv11_fuse}{\includegraphics[width=0.24\textwidth]{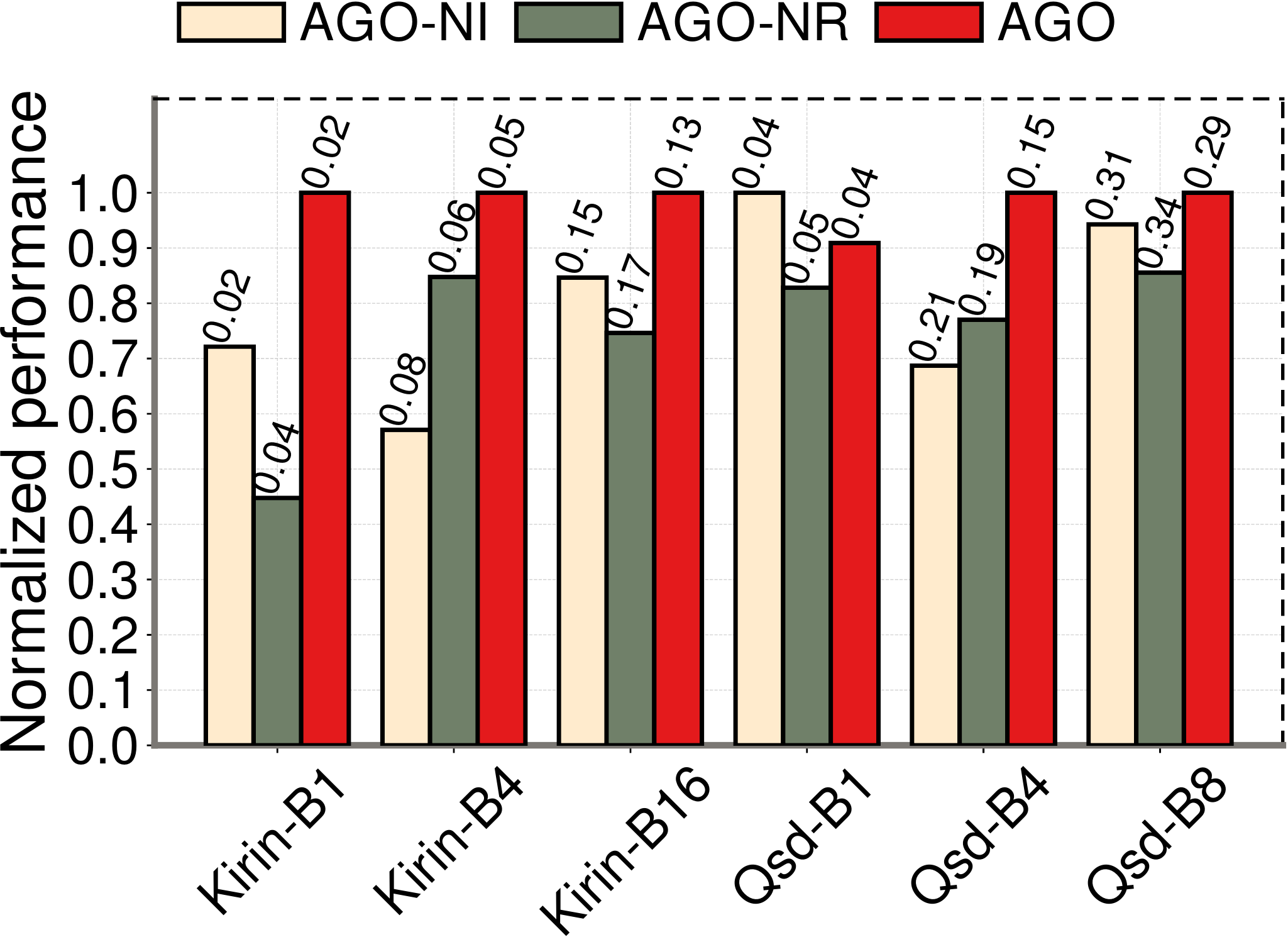}}}\hfill
	~
	\subfigure[Pointwise conv \& depthwise conv.]{\label{subfig:conv11_dep33_fuse}{\includegraphics[width=0.24\textwidth]{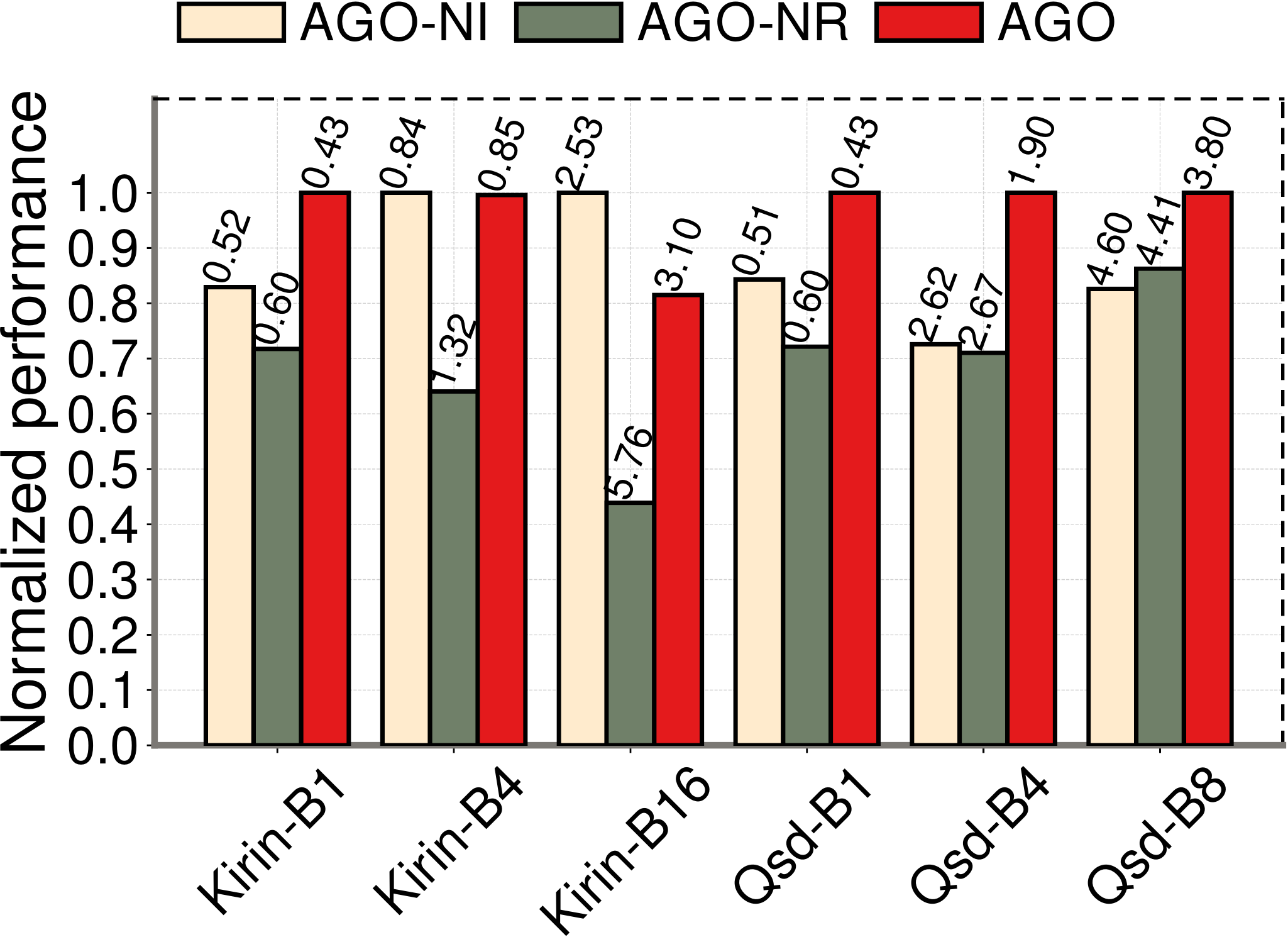}}}\hfill
	~
	\subfigure[Two pointwise convolutions.]{\label{subfig:conv11_conv11_fuse}{\includegraphics[width=0.24\textwidth]{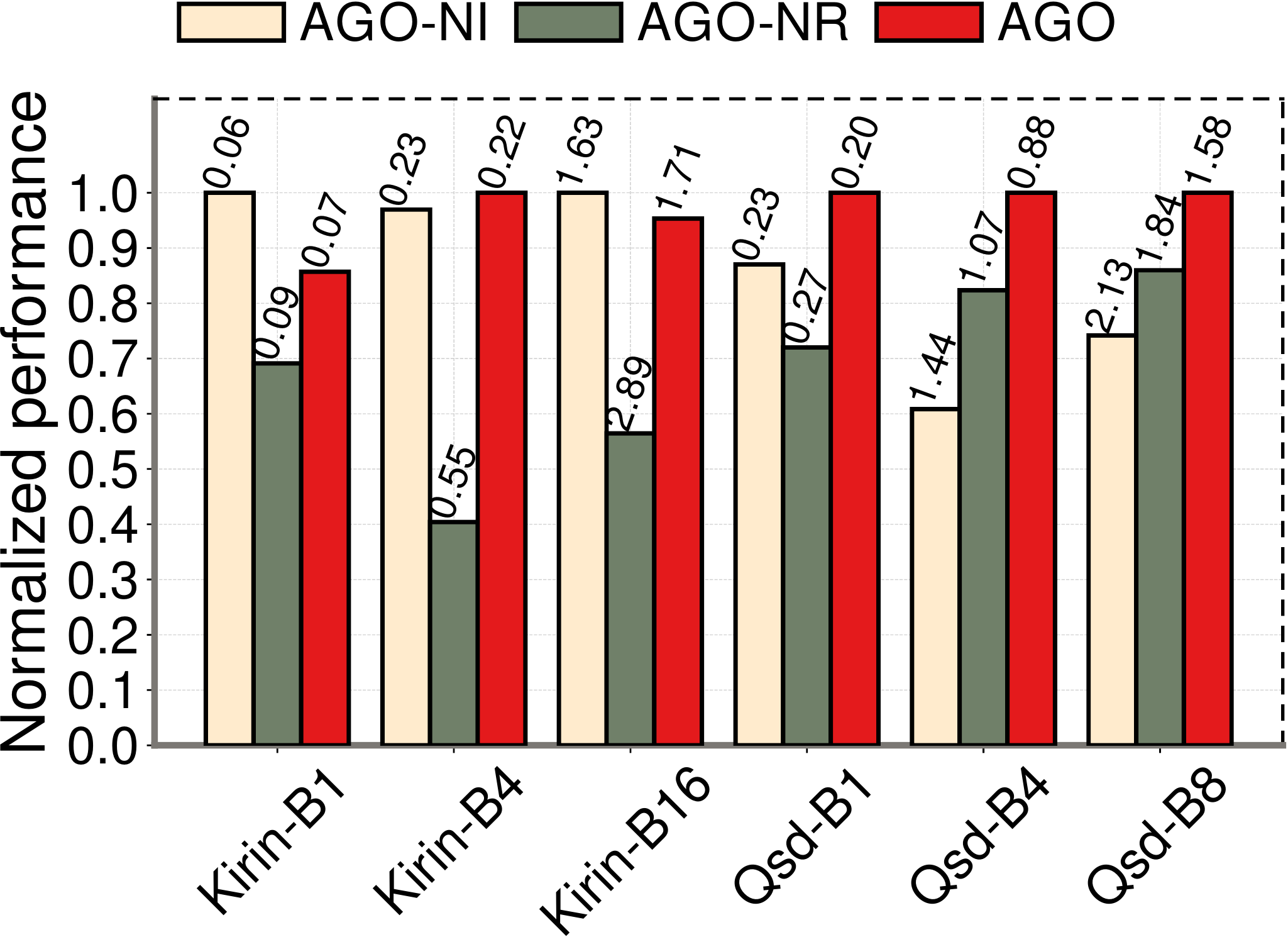}}}\hfill
	\vspace{-0.1in}
	\caption{Subgraph performance with different variants of \sys.}
	\label{fig:fusionstrategies}
	\vspace{-0.2in}
\end{figure*}

In this subsection, we further study where the performance gain of \sys comes from, to evaluate our intensive fusion and reformer layer.
Then, we will evaluate our graph partitioning algorithm, by respectively inspecting the generated subgraphs partitioned by our algorithm and Relay \cite{roesch2018relay}.

We first compare three variants of \sys to break down the improvements: 1) \sys-NI (no intensive fusion in the tuner backend); 2) \sys-NR (no reformer layer, \textit{i.e.}, tuning a large subgraph directly); 3) \sys, same as \cref{subsec:e2e}, as the baseline. 
We then evaluate them on four subgraphs. Each subgraph consists of two complex operators and some other simple operators. 
The complex operator is either pointwise convolution or depthwise convolution. 
Except the subgraph with two depthwise convolutions, other subgraphs are extracted from MBN and MNSN.
Additionally, the tuning budget is 2,000 for each variant and subgraph.
%
%

We present the results in \cref{fig:fusionstrategies}, where the number behind $B$ is the batch size.
\sys-NI has average $17\%$ performance loss compared with \sys on two platforms. 
The reason is that \sys can fuse multiple complex operators to further improve the performance, while \sys-NI only optimizes them jointly with conventional fusion.
Further, \sys-NR has nearly 27\% performance loss.
This is because directly optimizing a complicated subgraph is hard, while \sys addresses this issue through the divide-and-conquer tuning mechanism.
We also observe that there are some cases where \sys-NI outperforms \sys \cref{subfig:conv11_conv11_fuse}. 
This indicates that \sys cannot find better schedules due to the increased search space size after joining mini-subgraphs, hence inefficient budget usage in such workloads. 
Thus, this issue can be addressed by prolonging the tuning for mini-subgraphs before joining.

%

\begin{figure}
	\centering
	\includegraphics[width=0.46\textwidth]{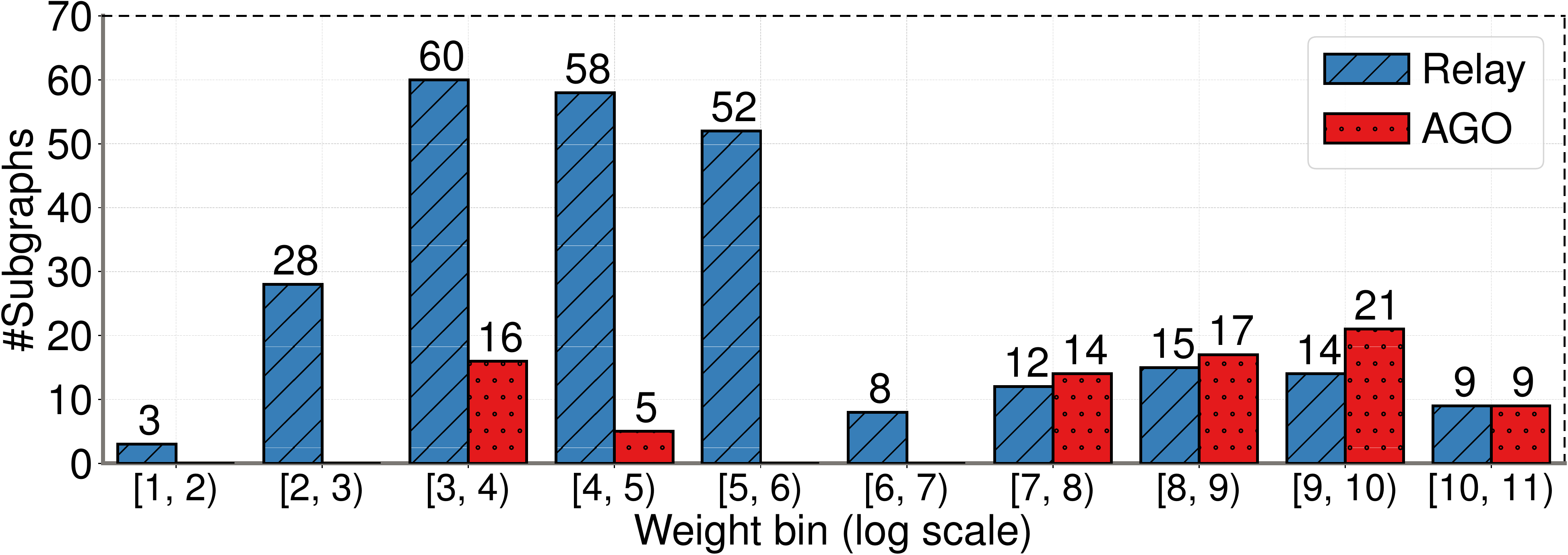}
	\vspace{-0.1in}
	\caption{Subgraph weight distribution for MVT.}
	\label{fig:weight_dist}
	\vspace{-0.2in}
\end{figure}

Next, we evaluate our graph partitioning algorithm.
We present the subgraph weight distribution for MVT in \cref{fig:weight_dist}. We construct ten weight bins in log scale (\eg, bin $[1, 2)$ means weight interval $[2^1, 2^2)$), and report the number of subgraphs in each bin.
The new model MVT integrates attention modules, yielding a large number of reshape and transpose operators. Relay will heuristically take such operators as delimiters to produce totally 259 subgraphs, where 105 of them are trivial and have a weight less than 20. Besides, the average weight, the median weight, and the Jain's fairness index (measuring balance and higher is better) are 138, 23, and 0.19, respectively.
In contrast, \sys generates 82 subgraphs. Most of them have a large weight as shown in \cref{fig:weight_dist}. For \sys, the average weight, the median weight, and the Jain index are 437, 350, and 0.55, respectively. Therefore, our partitioning algorithm can generate more complicated subgraphs while maintaining balance.
%
Take a typical structure in MVT as an example, which contains eight consecutive operators: matrix multiplication, reshape, add, reshape, transpose, reshape, matrix multiplication, and reshape.
Relay produces five fragmented subgraphs for this structure, missing opportunities of intensive fusion for two matrix multiplications and joint optimization for all simple operators. 
Such partition leads to inferior performance since the reshape/transpose operators involve expensive memory loads/stores.
By contrast, \sys typically groups such operators together to boost the performance. 

\section{Related Work}
\label{sec:related}
\noindent\textbf{Deep learning compiler:}
Various deep compilers have been proposed to address the error-prone manual optimization \cite{leary2017xla,chen2018tvm,chen2018learning,vasilache2018tensor,jia2019taso,zheng2020flextensor,zheng2020ansor,zheng2020fusionstitching,wang2021tuna,li2021analytical,steiner2021value,baghdadi2021deep,yang2021equality,akg,2021pet, xu2022alt}. 
Tensor Comprehension \cite{vasilache2018tensor} and TVM \cite{chen2018tvm} adopt the idea of decoupling optimization from operator description to simplify the auto-tuning process, which is then provided by AutoTVM \cite{chen2018learning}, FlexTensor \cite{zheng2020flextensor},  Ansor \cite{zheng2020ansor}, ALT \cite{xu2022alt} via search algorithms. TASO \cite{jia2019taso}, Tensat \cite{yang2021equality}, and PET \cite{2021pet} perform graph substitutions to generate more efficient graphs. To speed up the tuning, AKG \cite{akg} applies the polyhedron model, while delicate cost models \cite{wang2021tuna,li2021analytical} and heuristics \cite{steiner2021value, zhu2022roller} are also proposed. Compared with \sysname{}, 1) their tuners do not support intensive fusion; 2) their frontends take the tuner as a black box and no cross-layer mechanism is involved;
3) their graph partitioning algorithms impose unnecessary constraints on subgraph structures, thus can only generate simple subgraphs.

\noindent\textbf{Operator fusion:}
%
Many systems exploit operator fusion as an important optimization technique \cite{zheng2020fusionstitching, 2021DNNFusion, apollo,2021Bolt,akg, zheng2022astitch}. 
In general, they can fuse a complex operator with its following simple operators, which is named as conventional fusion in this work.
For instance, \cite{zheng2022astitch} can fuse memory-bound operators for NVIDIA GPU based on XLA \cite{leary2017xla}. 
\cite{akg, apollo} exploits the polyhedral model to explore the fusion opportunity.
Although two matrix multiplications can also be fused on NVIDIA GPU in Bolt \cite{2021Bolt}, it is implemented based on the vendor library CUTLASS \cite{cutlass} and the fact that the cache (shared memory) in NVIDIA GPU is programmable.
Thus, Bolt can only fuse matrix multiplications that CUTLASS supports, while cannot fuse general complex operators on CPUs.
Compared with these works, \sys enables generic auto-tuned intensive fusion on mobile devices, while guaranteeing efficiency via careful analysis on computation redundancy.
%
%
%

\noindent\textbf{Computational graph partitioning:}
Classical graph partitioning has been extensively studied \cite{malewicz2010pregel}, typically in distributed computing area.
In deep learning systems \cite{ding2021ios,ahn2020ordering,laskaridis2020spinn,huang2020clio, unity}, they are often used to increase the parallelism to improve performance. 
For instance, IOS \cite{ding2021ios} and Unity \cite{unity} exploit partitioning to parallelize subgraphs on NVIDIA GPU. 
\cite{ahn2020ordering} partitions the graph to reduce the peak runtime memory footprint. 
SPINN \cite{laskaridis2020spinn}, CLIO \cite{huang2020clio}, and Walle \cite{walle} partition a computational graph into two parts, with one part running on the device and the other running on the cloud/edge server.
%
Compared with these systems, the major goal of our graph partitioning is to improve mobile inference performance via compilation techniques while keeping acyclic theoretically.
%
\presec \section{Conclusion}\label{sec:conclusion}\postsec
We propose \sys, a framework that removes the constraints on graph optimization to boost the inference performance of AI models on mobile devices. 
\sys provides a new partitioning scheme to generate arbitrary subgraphs while keeping acyclic. 
It also designs a potent tuner which proposes intensive operator fusion and joint optimization to boost arbitrary subgraphs.
Additionally, \sys devises a divide-and-conquer mechanism to address the tuning efficiency. 
Experiments show that \sys significantly outperforms state-of-the-art hand-tuned libraries and makes great progress over auto-tuning frameworks. 
For researchers, they can use \sys to further explore the performance characteristics of complicated subgraphs. For practitioners, they can easily exploit \sys to improve the inference performance without human interference. 
\postsec
\section{Acknowledgement}
We thank the anonymous reviewers of INFOCOM 2023 for their valuable comments. This work is partially supported by the
National Natural Science Foundation of China under Grant
Number 62272213 and the Jiangsu Innovation and Entrepreneurship (Shuangchuang) Program.

}
\bibliographystyle{IEEEtran}
\bibliography{main}
\end{document}